\PassOptionsToPackage{table,xcdraw,dvipsnames}{xcolor}

\documentclass[numbers]{article}

\usepackage[final]{neurips_2024}
\usepackage{graphicx}
\usepackage{amsmath}
\usepackage{wrapfig}
\usepackage{float}
\usepackage{listings}
\usepackage{authblk}

\usepackage{multirow}
\usepackage{subcaption}
\usepackage{geometry}
\usepackage{caption}

\usepackage[utf8]{inputenc} 
\usepackage[T1]{fontenc}    
\usepackage{hyperref}       
\usepackage{url}            
\usepackage{booktabs}       
\usepackage{amsfonts}       
\usepackage{nicefrac}       
\usepackage{microtype}      
\usepackage{dblfloatfix}
\usepackage{pifont}
\usepackage{lineno}
\usepackage{hyperref}
\usepackage{url}
\usepackage{microtype}
\usepackage{hyperref}
\usepackage{url}
\usepackage{booktabs}
\usepackage{booktabs}
\usepackage{subcaption}
\usepackage{booktabs}
\usepackage{subcaption}
\usepackage{tabularx}
\usepackage{makecell}
\usepackage{lineno}
\usepackage{booktabs}
\usepackage{pifont}
\usepackage[utf8]{inputenc}
\usepackage{enumitem}
\usepackage{framed}
\usepackage{amsmath}
\usepackage[symbol]{footmisc}
\newcommand{\cmark}{\ding{51}}
\newcommand{\xmark}{\ding{55}}
\usepackage{soul}        

\usepackage[dvipsnames]{xcolor}

\sethlcolor{yellow}
\newcommand{\hlyellow}[1]{\hl{#1}}

\newcommand{\hlpurple}[1]{{%
  \sethlcolor{Purple}
  \hl{#1}%
}}

\definecolor{darkblue}{rgb}{0, 0, 0.5}
\hypersetup{colorlinks=true, citecolor=darkblue, linkcolor=darkblue, urlcolor=darkblue}

\title{Evaluating Multi-Hop Reasoning in Large Language Models: A Chemistry-Centric Case Study}

\author{
\begin{minipage}[c]{\textwidth}
\centering
\textbf{
  Mohammad Khodadad\textsuperscript{1,2,†}, 
  Ali Shiraee Kasmaee\textsuperscript{1,2,*,†}, 
  Mahdi Astaraki \textsuperscript{1,2}, 
  Nicholas Sherck\textsuperscript{3}, 
  Hamidreza Mahyar\textsuperscript{1}, 
  Soheila Samiee\textsuperscript{2}}\\
  \textsuperscript{1}Department of Computational Science and Engineering, McMaster University, Canada \\
  \vspace{0.2em}
  \textsuperscript{2}BASF Canada Inc., Canada\\
  \textsuperscript{3}BASF Corporation, USA\\
  \texttt{\{khodam3, shiraeea, astarakm, mahyarh\}@mcmaster.ca \linebreak
  \{nicholas.sherck, soheila.samiee\}@basf.com} \\
  \textsuperscript{*}Corresponding Author: \texttt{shiraeea@mcmaster.ca}\\
    \textsuperscript{†}Equal Contribution
\end{minipage}
}

\begin{document}
\maketitle

\begin{abstract}
In this study, we introduced a new benchmark consisting of a curated dataset and a defined evaluation process to assess the compositional reasoning capabilities of large language models within the chemistry domain. We designed and validated a fully automated pipeline, verified by subject matter experts, to facilitate this task. Our approach integrates OpenAI reasoning models with named entity recognition (NER) systems to extract chemical entities from recent literature, which are then augmented with external knowledge bases to form a comprehensive knowledge graph. By generating multi-hop questions across these graphs, we assess LLM performance in both context-augmented and non-context augmented settings. Our experiments reveal that even state-of-the-art models face significant challenges in multi-hop compositional reasoning. The results reflect the importance of augmenting LLMs with document retrieval, which can have a substantial impact on improving their performance. However, even perfect retrieval accuracy with full context does not eliminate reasoning errors, underscoring the complexity of compositional reasoning. This work not only benchmarks and highlights the limitations of current LLMs, but also presents a novel data generation pipeline capable of producing challenging reasoning datasets across various domains. Overall, this research advances our understanding of reasoning in computational linguistics.
\end{abstract}

\section{Introduction}
Large Language Models (LLMs) have achieved impressive performance on a wide range of tasks, yet their ability to perform complex, multi-step reasoning remains an ongoing challenge. Techniques such as chain-of-thought (CoT) prompting \cite{wei2022chain, wang2022self, yao2023tree, besta2024graph, xiang2025towards} and structural innovations \cite{lewis2020retrieval, d2020neurosymbolic, santoro2017simple} have enabled notable improvements in reasoning, particularly in mathematics and coding. OpenAI’s o-series \cite{openaiO1SystemCard, openaiO3MiniSystemCard} was among the first to introduce inference-time scaling of CoT reasoning depth, and subsequent open-source models such as DeepSeek R1 \cite{zelikman2022star, guo2025deepseek} and Qwen QwQ \cite{patil2025advancing} have adopted similar strategies. Notably, these models also leverage reinforcement learning during training to further refine their CoT reasoning, consistently improving performance with increased test-time compute.

To evaluate the reasoning capabilities of LLMs, the community relies on a suite of benchmarks requiring multi-hop reasoning, spanning domains from mathematics \cite{cobbe2021training, hendrycks2021measuring} and programming \cite{chen2021evaluating, austin2021program, jimenez2023swe} to general question answering, including open-book tasks such as HotpotQA \cite{yang2018hotpotqa} and StrategyQA \cite{geva2021did}. Recent advances in scaling reasoning have led to improvements in these benchmarks. However, these models remain largely general-purpose. In scientific fields such as chemistry, multi-hop reasoning is essential for integrating interconnected, domain-specific knowledge. Although several multi-hop question answering benchmarks exist, evaluations specific to chemical reasoning are limited \cite{wellawatte2024chemlit, huang2024olympicarena, rein2024gpqa}. Recent work, including datasets targeting subfields such as reticular chemistry \cite{zheng2025large}, highlights the need for more comprehensive and challenging domain-specific benchmarks.

To address this gap, we propose an automated multi-hop reasoning data generation pipeline that leverages OpenAI's \textbf{\texttt{o3-mini}} and \textbf{\texttt{gpt-4o}} models. Our pipeline systematically extracts and verifies chemical entities via named entity recognition (NER) and links them to external databases to construct a knowledge graph, from which challenging multi-hop question-answer pairs are generated. Our contributions are as follows: 

\begin{enumerate} 
\item We provide extensive experimental evidence that compositional reasoning in scientific domains remains a significant limitation for current state-of-the-art LLMs. 
\item We demonstrate that even retrieval augmentation with perfect context does not guarantee flawless reasoning due to inherent compositional complexities. 
\item We introduce an automated knowledge graph and multi-hop reasoning data generation pipeline, leveraging OpenAI models and NER, which can potentially generate unlimited domain-specific datasets. 
\item We contribute a challenging benchmark for Multi-hop QA in chemistry, reviewed by our domain experts. 
\end{enumerate}

\section{Background and related works}

\paragraph{Multi-hop Question Answering.}
Multi-hop question answering (QA) has evolved as a key method to evaluate the multi-step reasoning abilities of large language models \cite{mavi2024multi}. Answering multi-hop questions requires integrating multiple pieces of evidence.
Traditionally, datasets such as HotpotQA \cite{yang2018hotpotqa}, WikiHop, and MedHop \cite{welbl2018constructing} were manually or semi-automatically curated through crowd-sourcing and knowledge-base relations. While these approaches yield high-quality, human-validated questions, they are resource-intensive.
More recently, LLMs are leveraged to automatically generate multi-hop datasets. In many cases, single-hop QA pairs are first generated and later merged using entity linking techniques, a common approach that connects individual entities across questions. For example, the MuSiQue framework \cite{trivedi2022musique} fuses two QA pairs by linking a named entity from the first answer to the subsequent question, thereby forming a chain of reasoning. Other methods, such as MultiHop-RAG \cite{tang2024multihop}, extend this paradigm by incorporating retrieval-augmented generation (RAG) to paraphrase factual sentences and group them based on shared topics, reflecting the diverse strategies that are emerging in multi-hop QA generation.

\paragraph{Chemistry Domain.}
The chemical sciences pose distinct challenges for multi-hop QA due to the need for expert domain knowledge. Only a small fraction of HotpotQA’s questions (roughly 900) are chemistry-focused, limiting both domain relevance and topical currency. Furthermore, they tend to be limited to 2-hop reasoning, which limits the difficulty level of the questions. 
More specialized datasets, such as ChemLitQA \cite{wellawatte2024chemlit}, provide around 1,000 single-hop and 700 multi-hop question-answer pairs generated using LLMs based on ChemRxiv papers. Although the ChemLitQA-multi dataset includes multi-hop questions, these are typically defined by a single linked entity across all hops. This limitation has led the authors of \cite{wellawatte2024chemlit} to emphasize the need for future studies focused on developing more challenging multi-hop QA benchmarks in the field.

Similarly, the chemistry subset of the OlympicArena benchmark \cite{huang2024olympicarena} provides high-level Olympiad problems that are few in number, 
and are not explicitly designed for multi-hop reasoning. As another recent effort, \cite{mirza2025framework} introduces ChemBench, an automated benchmark of over 2,700 expert‑validated chemistry QA pairs to systematically assess LLMs’ chemical reasoning against human experts.

\begin{figure*}[!t]
    \centering
    \includegraphics[width=\textwidth]{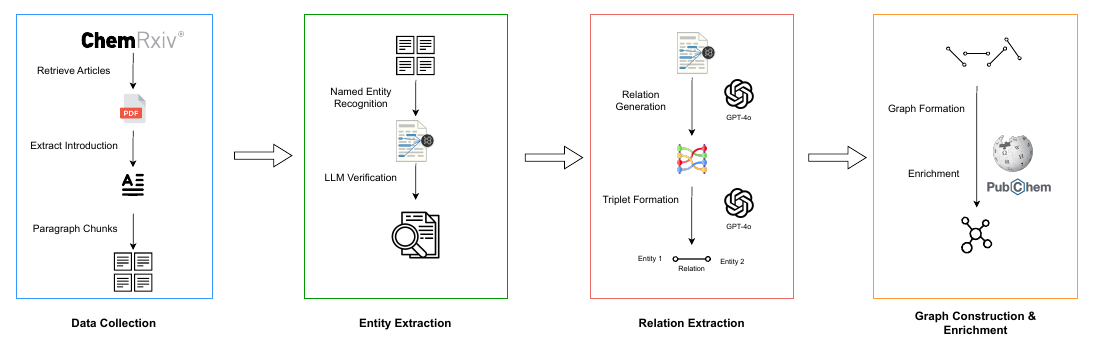}
    \caption{An Overview of the knowledge graph generation pipeline. }
    \label{fig:kg-generation}
\end{figure*}

\paragraph{Knowledge Graph Generation.}

Automated knowledge graph construction from unstructured text has seen diverse approaches aimed at effectively capturing entities and their interrelations. Early systems, such as Grapher \cite{melnyk2022knowledge}, take an end-to-end approach by first generating nodes with a fine-tuned pretrained language model and then forming edges via sequential generation or classification techniques. Building on these ideas, frameworks like the Extract-Define-Canonicalize (EDC) approach \cite{zhang2024extract} employ a three-phase strategy: they extract relational triplets without a fixed schema, generate natural language definitions for each relation, and then standardize and merge equivalent triplets, sometimes with an additional refinement stage that leverages retrieval techniques.

Complementing these general-purpose methods, dynamic and domain-specific approaches address specialized challenges in KG construction. For instance, KG-MRC \cite{das2018building} models the evolution of entity states in procedural texts by integrating neural reading comprehension with recurrent graph updates to capture temporal changes. In parallel, domain-specific systems such as CEAR \cite{langer2024cear} incorporate tailored ontologies and specialized knowledge to generate more accurate graphs from the scientific literature. Similarly, Cai et al. \cite{liao2023coarse} demonstrate an iterative, coarse-to-fine refinement process that adapts a broad biomedical knowledge graph to specialized domains like oncology, reducing reliance on manual annotations while preserving essential domain nuances.

\section{Methodology}
Our methodology consists of three main components: knowledge graph generation, multi-hop question-answer generation, and evaluating state-of-the-art large language models on the question-answering task. The first two components are described in this section, and the last one is explained in the next section.
\subsection{Knowledge Graph Generation}
We began by constructing a comprehensive knowledge graph from chemical literature. First, using the ChemRxiv API, we collected all ChemRxiv articles with licenses that permitted redistribution. Next, we cleaned the articles using regular expressions to extract their introductions. Focusing on objective and factual information, we extracted each introduction's first few paragraphs (up to 500 words). Finally, we segmented the extracted text into chunks of up to 128 words, ensuring that no paragraph was split across chunks.

Next, we applied named entity recognition (NER) models to these text chunks to identify chemical entities. In particular, we utilized an NER model \cite{ruas2022nilinker} that leverages a PubMedBERT architecture \cite{pubmedbert} fine-tuned on various chemical datasets. To ensure that the extracted entities were specific, verified, and chemically relevant, we utilized OpenAI's \textbf{\texttt{gpt-4o}} to review and refine the outputs. The same model was also employed to extract relations between these verified entities, forming triplets that capture the interactions and associations present in the text. Additionally, large language models were utilized to extract descriptive features from textual data associated with each entity. To enrich the nodes further during the construction of the knowledge graph, supplementary information from Wikipedia and the PubChem dataset \cite{kim2021pubchem} was integrated. Consequently, the finalized knowledge graph comprises nodes representing chemical entities, enhanced by metadata and descriptive annotations from these sources, as well as edges representing the relationships extracted from the textual data. Figure~\ref{fig:kg-generation} illustrates the procedure followed to generate the knowledge graph. Refer to the Appendix for a detailed explanation of Knowledge Graph Generation.

\begin{figure*}[!t]
    \centering
    \includegraphics[width=\textwidth]{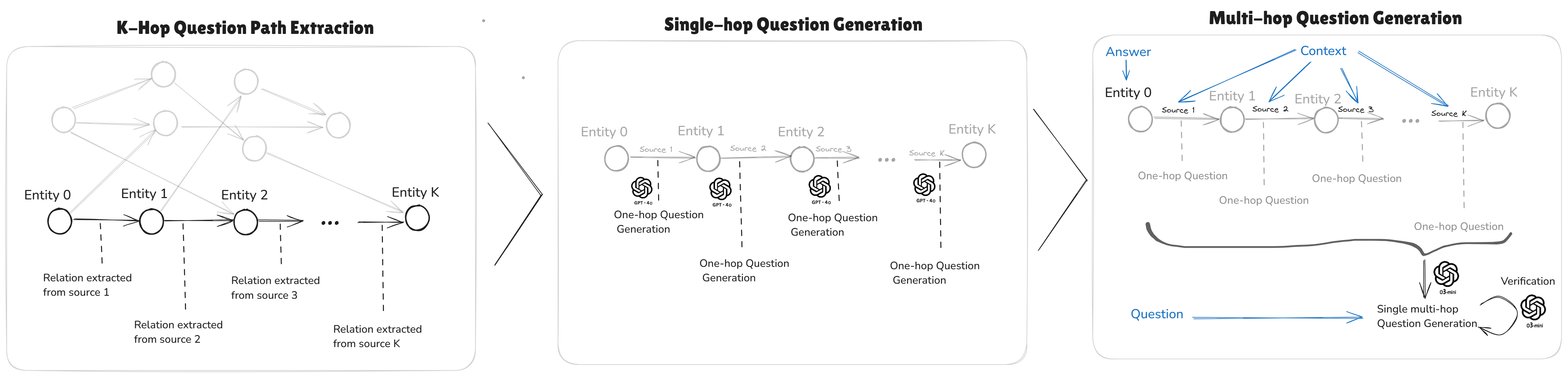}
    \caption{Overview of the QA generation Pipeline.}
    \label{fig:qa_pipeline}
\end{figure*}

\subsection{Multi-hop Question-Answer Generation}
\label{sec:qa-generation}
To generate multi-hop questions, we first sampled paths of varying lengths from the constructed knowledge graph using a randomized breadth-first search (BFS) path sampling algorithm. During path sampling, we ensured that the sources for the edges were distinct, encouraging solutions to integrate information from multiple sources and different parts of the context to answer the questions. Therefore, each path with a length of K involves K+1 entities, coming from K distinct source texts extracted from the original ChemRxiv database.

Adopting a bottom-up approach, we began by sampling paths and generating individual 1-hop questions from each hop. Specifically, For every $(\text{entity}_1, \text{relation}, \text{entity}_2)$ triplet, we formulated a corresponding question in which $\text{entity}_1$ served as the answer, the prompt inquired about the entity that holds the specified relation to $\text{entity}_2$. When a question lacked sufficient specificity, we instructed an LLM to enrich it with additional metadata or context from the original text, thereby enhancing clarity and precision.

These individual questions were then combined into a single multi-hop question using OpenAI's \textbf{\texttt{o3-mini}} model. Importantly, the final aggregated question was constructed to begin with the last sub-question and chain the entities up to the $\text{entity}_1$ of the first relation, ensuring that the final answer corresponds to the answer of the first question. 

During the verification phase, each one-hop question was first reviewed for clarity, relevance to chemistry, and alignment with the corresponding text that provided the answer. The multi-hop question was then assessed through an additional evaluation step, ensuring that its logical flow effectively led to the final answer. An LLM-based verification process was employed to confirm factual accuracy, answerability based on available context and metadata, and the logical coherence of the sub-questions. Feedback from domain experts was continuously incorporated into the prompts to enhance verification accuracy.
To enhance the quality of the dataset, an additional round of verification was conducted on the final questions, resulting in the removal of a subset. Supplementary Section~\ref{subsec:rejected_q} provides more details on the types of questions that were dropped and the reasoning behind their exclusion. 
Figure~\ref{fig:qa_pipeline} illustrates the detailed pipeline of Multi-hop QA generation. Refer to the Appendix for a detailed explanation of QA Generation.

To minimize the impact of writing style and summarization on accuracy evaluation, all questions are designed to have short answers. Answering these questions requires breaking down the main question into smaller sub-questions, finding the answer to each, and combining them to arrive at the final answer. Even with full context available,  a correct answer cannot be obtained if the model is not capable of inferring and integrating different pieces of knowledge.

While our pipeline is largely automated, domain expert input was integral throughout development. We conducted two rounds of pilot generations followed by detailed reviews from chemists, using their feedback to improve the pipeline and significantly enhance the quality of the generated questions. Furthermore, as explained in Section~\ref{subsec:feedback}, a random subset of questions from the final dataset was evaluated by a domain expert to verify the quality of the generated dataset.

\subsection{Statistical Details of the Generated Graph and Dataset}

Table \ref{tab:combined_stats} provides a concise overview of both our dataset‐level and graph‐level statistics. In Table \ref{tab:dataset_stats}, we summarize key properties of the 971 multi‐hop questions, including average question and answer lengths (in characters and tokens), the mean number of hops per question, total and pooled context lengths, and the proportion of questions containing at least one shortcut edge. 

On average, questions were 319.4 chars long (SD = 129.2) or 45.5 tokens (SD = 18.6), while answers averaged just 16.7 chars (SD = 9.7) or 1.76 tokens (SD = 0.96). Each question required a mean of 2.45 hops (SD = 1.12). Summing all hops per question gives a total context length of $5,993$ chars (SD = $5,009$) or 849 tokens (SD = 726), and pooled hop lengths average $2,447$ chars (SD = $2,222$) and 346 tokens (SD = 324). Only 96 questions (9.9 \%) include at least one shortcut edge (mean = 0.12, SD = 0.38). The full hop-count breakdown appears in the "Hop-count Distribution" block of Table \ref{tab:dataset_stats}. 

\begin{table}[b!]
  \centering
  \scriptsize
  \setlength{\tabcolsep}{3pt}
  \renewcommand{\arraystretch}{0.95}

  \begin{subtable}[t]{0.48\textwidth}
    \centering
    \begin{tabularx}{\linewidth}{@{}X r r@{}}
      \toprule
      \textbf{QA Metric}                       & \textbf{Mean}   & \textbf{Std.\ Dev.} \\
      \midrule
      Question length (chars)               & 319.42          & 129.17             \\
      Question length (tokens)              & 45.49           & 18.64              \\
      Answer length (chars)                 & 16.66           & 9.69               \\
      Answer length (tokens)                & 1.76            & 0.96               \\
      Mean \# hops per question             & 2.45            & 1.12               \\
      Total context length (chars)          & 5993.10         & 5009.69            \\
      Total context length (tokens)         & 848.55          & 725.78             \\
      Hop length (chars, pooled)            & 2447.14         & 2222.35            \\
      Hop length (tokens, pooled)           & 346.49          & 324.56             \\
      Shortcut count per question           & 0.12            & 0.38               \\
      \midrule
      \multicolumn{3}{@{}l}{\textbf{Hop‐count Distribution (of 971 questions)}} \\
      1 hop                                  & \multicolumn{2}{r}{258 (26.6\%)}    \\
      2 hops                                 & \multicolumn{2}{r}{245 (25.2\%)}    \\
      3 hops                                 & \multicolumn{2}{r}{242 (24.9\%)}    \\
      4 hops                                 & \multicolumn{2}{r}{226 (23.3\%)}    \\
      $\ge$ 5 hops                                & \multicolumn{2}{r}{0   (0\%)}       \\
      \midrule
      \textbf{Questions w/ $\ge$ 1 shortcut}      & \multicolumn{2}{r}{96 (9.9\%)}     \\
      \bottomrule
    \end{tabularx}
    \caption{Dataset‐level statistics for multi‐hop questions}
    \label{tab:dataset_stats}
  \end{subtable}%
  \hfill
  \begin{subtable}[t]{0.48\textwidth}
    \centering
    \begin{tabularx}{\linewidth}{@{}X r@{}}
      \toprule
      \textbf{Graph Metric}                & \textbf{Value}     \\
      \midrule
      Number of nodes                      & 14\,523            \\
      Number of edges                      & 13\,419            \\
      Density                              & 0.000127           \\
      Degree (min / max / avg)             & 0 / 257 / 1.85     \\
      Connected components                 & 4\,684             \\
      Largest component size               & 7\,318             \\
      Avg.\ clustering coefficient         & 0.0298             \\
      Degree assortativity coefficient     & –0.0265            \\
      \midrule
      \multicolumn{2}{@{}l}{\textbf{Top 5 nodes by degree}}                                 \\
      \multicolumn{2}{@{}l}{
        \makecell[tl]{hydrogen (257), carbon (250), oxygen (232),\\
                       CO\textsubscript{2} (220), lithium (155)}
      }                                                                               \\
      \bottomrule
    \end{tabularx}
    \caption{Key network‐level properties of the loaded knowledge graph}
    \label{tab:graph_stats}
  \end{subtable}

  \caption{Overview of both dataset‐level and graph‐level statistics. Left panel: Table \ref{tab:dataset_stats} summarizes the dataset‐level statistics for our 971 multi-hop questions. 
  Right panel: Key properties of the underlying knowledge graph are reported in Table \ref{tab:graph_stats}. 
  }
  \label{tab:combined_stats}
\end{table}

Table \ref{tab:graph_stats} then reports the main network characteristics of the underlying knowledge graph: its size (nodes and edges), sparsity (density), degree distribution (min, max, and average), number of connected components and the size of the largest component, as well as clustering and assortativity coefficients. 
The graph contains $14,523$ nodes and $13,419$ edges (density = 0.000127), with node degrees ranging from 0 to 257 (mean = 1.85). It splits into 4,684 connected components, the largest of which spans $7,318$ nodes. The average clustering coefficient is 0.0298, and the degree assortativity coefficient is –0.0265.
Finally, the five highest‐degree nodes, hydrogen, carbon, oxygen, CO\textsubscript{2}, and lithium, are listed to highlight the most central concepts in the graph.

\noindent Table~\ref{tab:compact_comparison} compares our chemical dataset (ChemKGMultiHopQA) with HotpotQA-Chemistry and ChemLitQA-multi across question count, bridged entities, entity types, answer format, domain, and source. ChemKGMultiHopQA comprises 971 ChemRxiv questions enhanced by PubChem and Wikipedia (1–4 hops, auto-built KG with expert-verified subset), offering richer multi-hop chemical reasoning than HotpotQA-Chemistry (980 Wikipedia questions, 2 hops, no chemical entities) and ChemLitQA-multi (742 ChemRxiv questions, 1 entity, LLM+expert verification).

\begin{table*}[t!]
  
  \resizebox{\textwidth}{!}{%

    \centering
    \begin{tabular}{@{}l c c l l l l}
      \toprule
      \textbf{Dataset} & \textbf{\# Qs} & \textbf{\# bridged entities} &  \textbf{\# entity type} &\textbf{Answer type} & \textbf{Domain} & \textbf{Source} \\
      \midrule
      HotpotQA-Chemistry            & 980 & 2 & General  & Short & Wikipedia                & Crowd (Wiki)                    \\
      ChemLitQA-multi     & 742    & 1 & Chemistry  &  Long \& Short & ChemRxiv        & LLM + expert verified           \\
      ChemKGMultiHopQA    & 971    & 1–4 & Chemistry & Short & ChemRxiv enhanced by PubChem \& Wikipedia     & LLM + NER + KG (auto) + An expert verified subset          \\
      \bottomrule
    \end{tabular}
    }
    \caption{Comparison of HotpotQA, ChemLitQA-multi, and ChemKGMultiHopQA data sets.}
    \label{tab:compact_comparison}
\end{table*}

\section{Experiments and Results}
In our experiments, we evaluated the domain-specific multi-hop question-answering capabilities of a wide variety of state-of-the-art large language models, including both reasoning-focused and general-purpose models. For clarity, throughout this work we refer to models specifically optimized to scale test-time compute as \textbf{reasoning models}. These included open-source and proprietary variants, tested with or without provided context. The summary of tested models and their performance is provided in Table~\ref{tab:performance_full}. 

\begin{table*}[b!]
  \centering
  \resizebox{\textwidth}{!}{%
    \begin{tabular}{lccccccc}
      \toprule
      Model & Context & Correctness Rate (\%) & Avg Duration (s) & Avg Input Tokens & Avg Output Tokens & Total Input Tokens (K) & Total Output Tokens (K) \\
      \midrule

        Anthropic Claude Sonnet 3.5 V2 & \xmark & 40.06 & 1.54 & 567 & 29 & 550.93 & 28.69 \\
        Anthropic Claude Sonnet 3.5 V2 & \cmark & 72.50 & 1.68 & 2210 & 30 & 2146.11 & 29.18 \\
        Anthropic Claude Sonnet 3.7 & \xmark & 44.80 & 1.61 & 567 & 30 & 550.93 & 29.35 \\
        Anthropic Claude Sonnet 3.7 & \cmark & 80.02 & 1.84 & 2210 & 30 & 2146.11 & 29.49 \\
        Anthropic Claude Sonnet 3.7 (Thinking) & \xmark & 45.73 & 39.01 & 583 & 1777 & 566.09 & 1725.79 \\
        Anthropic Claude Sonnet 3.7 (Thinking) & \cmark & 84.35 & 15.35 & 2228 & 715 & 2163.59 & 694.78 \\
        OpenAI GPT-4o-mini & \xmark & 32.34 & 0.63 & 204 & 9 & 198.60 & 9.63 \\
        OpenAI GPT-4o-mini & \cmark & 62.82 & 0.71 & 1628 & 10 & 1581.57 & 10.01 \\
        OpenAI GPT-4o & \xmark & 40.27 & 0.63 & 204 & 9 & 198.60 & 9.53 \\
        OpenAI GPT-4o & \cmark & 68.80 & 0.71 & 1628 & 10 & 1581.57 & 9.95 \\
        OpenAI o1-mini & \xmark & 41.09 & 7.78 & 160 & 1047 & 155.88 & 1017.55 \\
        OpenAI o1-mini & \cmark & 71.99 & 5.68 & 1609 & 718 & 1562.70 & 697.41 \\
        OpenAI o3-mini & \xmark & 47.58 & 10.84 & 210 & 1187 & 204.43 & 1153.12 \\
        OpenAI o3-mini & \cmark & 80.33 & 6.12 & 1634 & 558 & 1587.40 & 542.46 \\
        Mistral Large & \xmark & 35.53 & 0.41 & 177 & 13 & 172.45 & 13.40 \\
        Mistral Large & \cmark & 73.94 & 0.57 & 1913 & 14 & 1857.70 & 14.22 \\
        Llama 3.3 70B Instruct & \xmark & 32.13 & 0.33 & 330 & 10 & 320.47 & 10.56 \\
        Llama 3.3 70B Instruct & \cmark & 65.19 & 0.40 & 1781 & 11 & 1729.91 & 10.75 \\
        Google Gemma 3 27B & \xmark & 32.03 & 0.89 & 163 & 12 & 158.95 & 11.94 \\
        Google Gemma 3 27B & \cmark & 69.72 & 1.00 & 1587 & 12 & 1541.55 & 12.57 \\
        DeepSeek R1 & \xmark & 44.39 & 21.06 & 159 & 1466 & 154.40 & 1423.73 \\
        DeepSeek R1 & \cmark & 81.98 & 8.61 & 1551 & 573 & 1506.14 & 556.55 \\
        Qwen QwQ 32B & \xmark & 35.74 & 68.29 & 168 & 2167 & 163.51 & 2104.86 \\
        Qwen QwQ 32B & \cmark & 79.81 & 25.18 & 1665 & 757 & 1617.45 & 735.86 \\
        DeepSeek R1 Distill Qwen 32B & \xmark & 34.19 & 32.04 & 159 & 1074 & 154.70 & 1043.56 \\
        DeepSeek R1 Distill Qwen 32B & \cmark & 79.09 & 12.11 & 1633 & 400 & 1586.25 & 389.31 \\
      \bottomrule
    \end{tabular}%
  }
  \caption{Summary of tested models performance in terms of several evaluation metrics for both Contextual and Non-Contextual Setups}
  \label{tab:performance_full}
\end{table*}

To access the selected models for evaluation in this experiment, we used different API providers: (i) all tested OpenAI models (\texttt{gpt-4o}, \texttt{gpt-4o-mini}, \texttt{o1-mini} and \texttt{o3-mini}) are accessed via the OpenAI platform; (ii) Amazon Bedrock Platform has been used to access \texttt{Anthropic Sonnet 3.7} (with and without extended thinking), \texttt{Anthropic Sonnet 3.5 V2}, \texttt{Mistral Large}, \texttt{DeepSeek R1} and \texttt{Llama 3.3 70B Instruct}; and (iii) \texttt{Google Gemma 3 27B}, \texttt{Qwen QwQ 32B}, and \texttt{DeepSeek R1 Distill Qwen 32B} are accessed via the OpenRouter Platform and operate at bf16 precision. All of these models can perform function-calling tool use, so they were instructed to produce valid JSON outputs to ensure consistency and enable automated validation. All models are evaluated in two settings: with and without provided context. The first scenario reflects performance when the models are paired with an ideal retrieval-augmented generation (RAG) system, while the second scenario relies on the model's internal memory to answer the questions. After parsing the JSON, we checked whether the output was an exact match to the ground truth. If not, OpenAI GPT-4o was instructed to perform a binary assessment, determining whether the answer was correct or not, to calculate the \textbf{Correctness Rate} (\%) metric. Our dataset comprises 971 questions spanning 1 to 4 hops (On average, 245 questions per hop), generated using the approach described in Section \ref{sec:qa-generation}.
The full Q\&A dataset, along with the evaluation code, is accessible \href{https://anonymous.4open.science/r/ChemMultiHop-6C47/}{here}.

\subsection{Models Performance}
\begin{figure*}[hbt!]
    \centering
    \includegraphics[width=.99\textwidth]{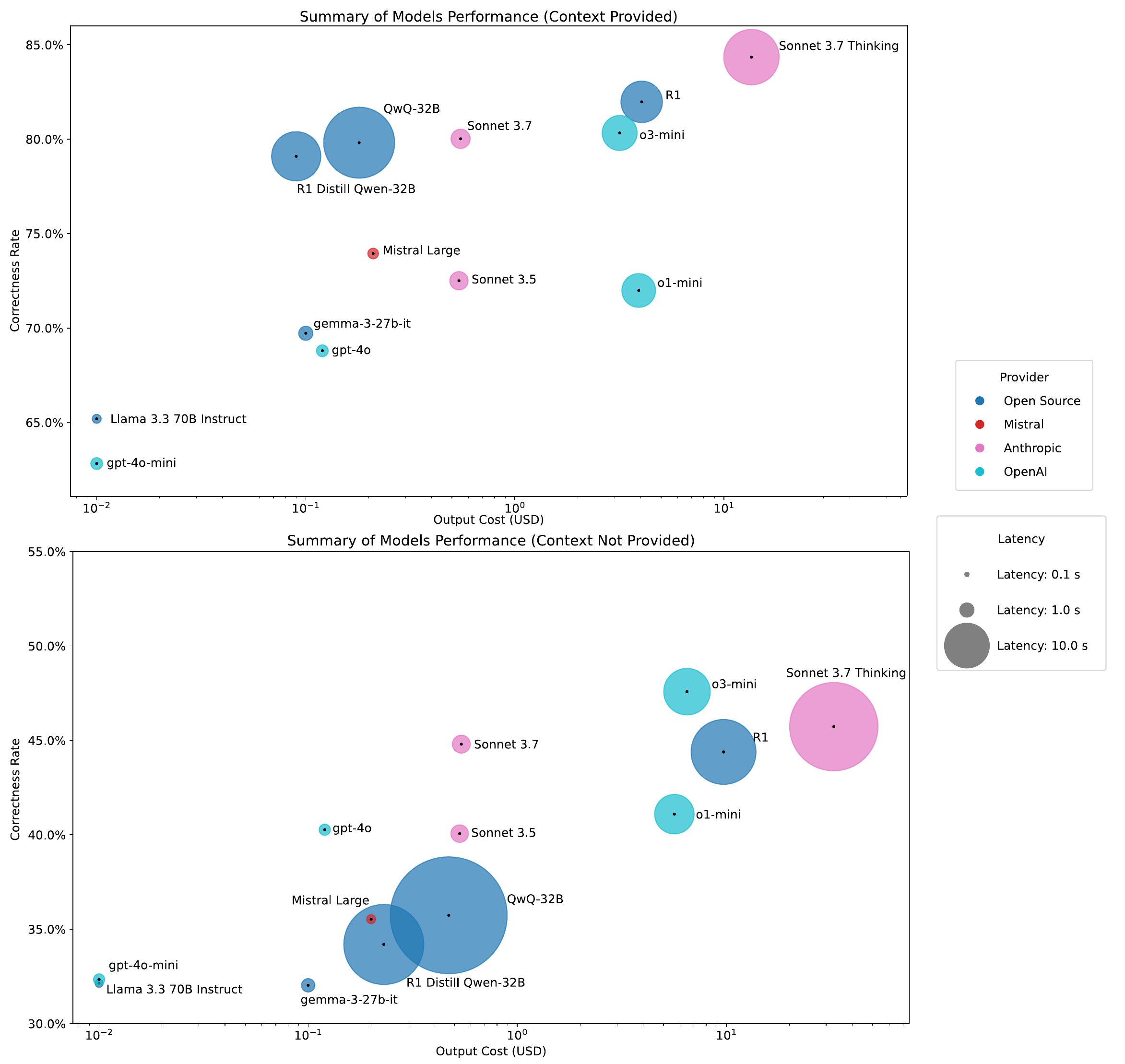}
    \caption{Performance of selected models based on correctness rate, cost, and latency. The cost axis is log-scaled to emphasize differences. The y-axis represents the percentage of questions answered correctly by each model, while the size of the dots indicates the average latency for each model when answering questions. The top panel displays results for setups where context is provided to the models, while the bottom panel presents results for setups without context. Note that the horizontal axis in both panels has the same range, while the vertical axes have different ranges.}
    \label{fig:performance}
\end{figure*}

Figure~\ref{fig:performance} illustrates the performance of 13 large language models evaluated with respect to correctness rate, cost, and latency in both \textit{context-provided} and \textit{context-not-provided} setups. 
In our performance evaluation, the \texttt{Llama 3.3 70B Instruct} and \texttt{GPT-4o} models achieved the lowest cost and demonstrated notably low latency, but they also registered the lowest correctness rate, making them a cost-efficient yet less accurate options. In contrast, Claude Sonnet 3.7 (with extended thinking) achieved the highest correctness rate, albeit at the expense of significantly higher cost and latency. Meanwhile, both Qwen QWEN 32B and Deepseek R1 Distil QWEN 32B maintained a favorable balance between cost and correctness rate when context is provided -- i.e., equipped with a perfect RAG system --, though they incurred above-average latency. Claude Sonnet 3.7 was found to have the highest correctness rate in \textit{none-reasoning} models. 

In \textit{Context not provided} setup, open-source reasoning models (\texttt{R1 Distill Qwen}, \texttt{QWQ-32B}, and \texttt{R1}) tend to have lower performance ranking compared to other models. In contrast, for OpenAI models, we observed an opposite trend, which may indicate potentially richer pre-training data. For Claude 3.7, using extended thinking did not lead to better performance compared to the no-thinking setup when the context is not provided to the model; instead, it increased the output tokens and cost. 
For a comprehensive breakdown of model performance and experimental settings, refer to Table~\ref{tab:performance_full}, which details metrics such as correctness rate, latency, and token usage. Note that the output details for reasoning models also include the reasoning tokens.

\subsection{Comparison with HotpotQA and ChemlitQA}

To demonstrate that large language models, even those designed for reasoning, often struggle with domain-specific multi-hop questions, we sampled a chemistry-related subset of HotpotQA \cite{yang2018hotpotqa}, a well-known general text benchmark primarily sourced from Wikipedia. We sampled chemistry questions by starting from Wikipedia's Chemistry category, recursively exploring its subcategories (up to three levels), and then filtering HotpotQA based on exact title matches. To maintain consistency with our evaluation scheme, we excluded distractors and included only supporting documents as context. This chemistry-specific subset of HotpotQA data is made available \href{https://anonymous.4open.science/r/ChemMultiHop-6C47/}{here}.

\begin{figure}[htb!]
    \centering
    \includegraphics[width=0.5\linewidth]{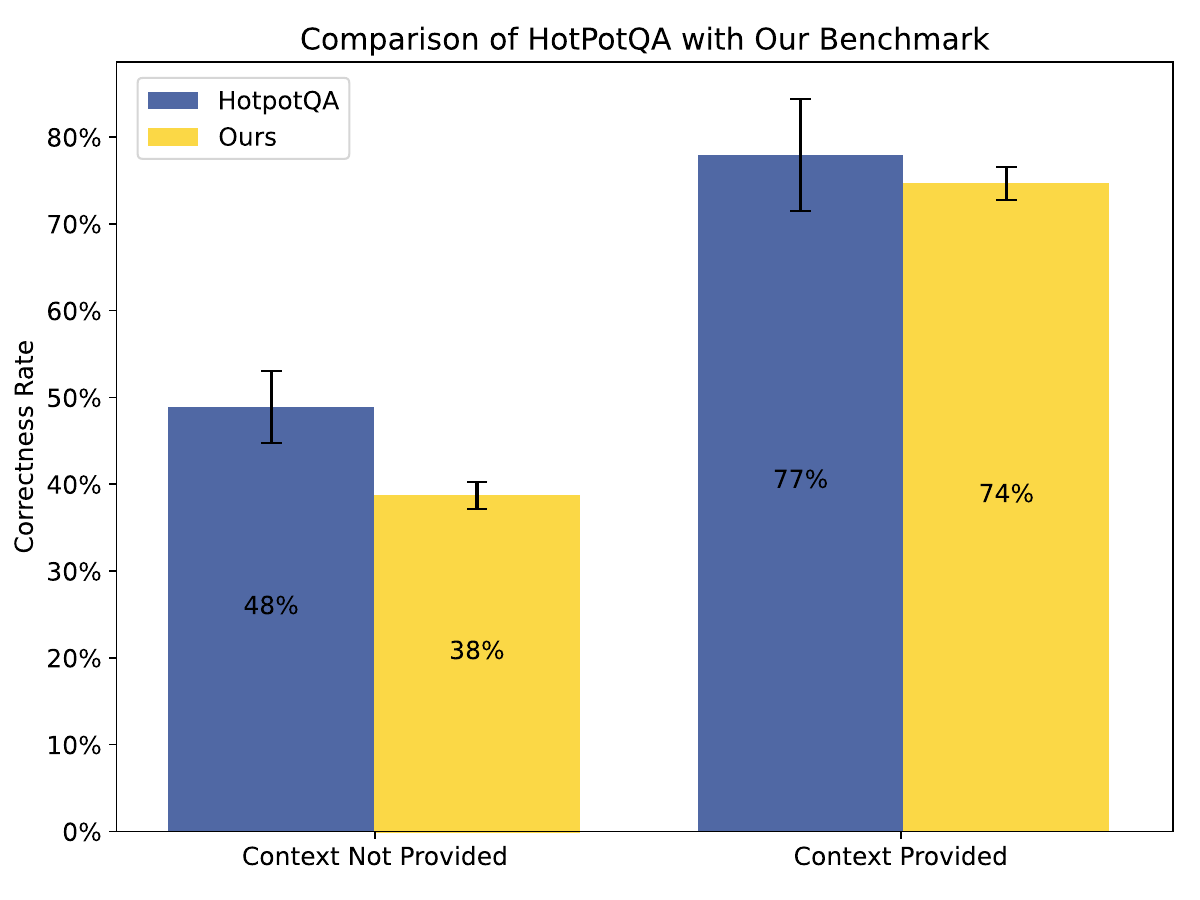}
    \caption{Comparison of LLMs' performance on the chemical subset of HotPotQA with the curated QA dataset from this study. 
    Error bars reflect the standard error of the mean (S.E.M) over evaluated models.}
    \label{fig:ours_vs_hotpotqa}
\end{figure}

Figure \ref{fig:ours_vs_hotpotqa} illustrates the average performance of all models on each dataset under two conditions: with context provided and without context in the prompt.
The results indicate that when context is provided, models achieve similar performance, with our benchmark resulting in marginally lower average performance and reduced variability among the models' output.
In the setup without context, the models found our benchmark more challenging compared to the HotpotQA chemistry subset. This observation may be due to the fact that HotpotQA was exclusively built from Wikipedia, which has been utilized in the pre-training of all evaluated models, while the new benchmark is constructed from more recent ChemRxiv papers enriched with PubChem and Wikipedia.

\section{Analysis and Ablation}
This section presents detailed ablation studies and analyses of the benchmark. First, we will present the results of a manual evaluation conducted by our panel of domain experts on a subset of the final dataset. Next, we examine how context availability and test-time reasoning affect model performance and efficiency. Finally, we will explore how the number of reasoning hops, an indicator of question difficulty, impacts model accuracy and the number of tokens needed to generate an answer.

\subsection{Expert Feedback}
\label{subsec:feedback}
A panel of domain experts with PhDs in Chemistry was invited to review a randomly selected subset of questions from the database. This review focused on the accuracy and quality of both the questions and their corresponding answers, as well as the necessity of multi-hop reasoning for answering them.
We started with 52 multi-hop questions, each paired with a fully worked, hop-by-hop answer.
Twelve questions (23 \%) were dropped due to low evaluator confidence, leaving 40 questions with high-confidence judgments. As shown in Supplementary Table \ref{tab:quality_metrics}, these 40 questions fall into three expert-rated categories: Good (26 questions, 65 \%), Ok (9 questions, 22.5 \%), and Poor (5 questions, 12.5 \%), generally approving 87.5\% of the evaluated questions. More detailed analysis is provided in  section~\ref{subsec:expert_feedback}.

\subsection{Context and Reasoning}
\begin{figure}[h]
    \centering
    \includegraphics[width=1\linewidth]{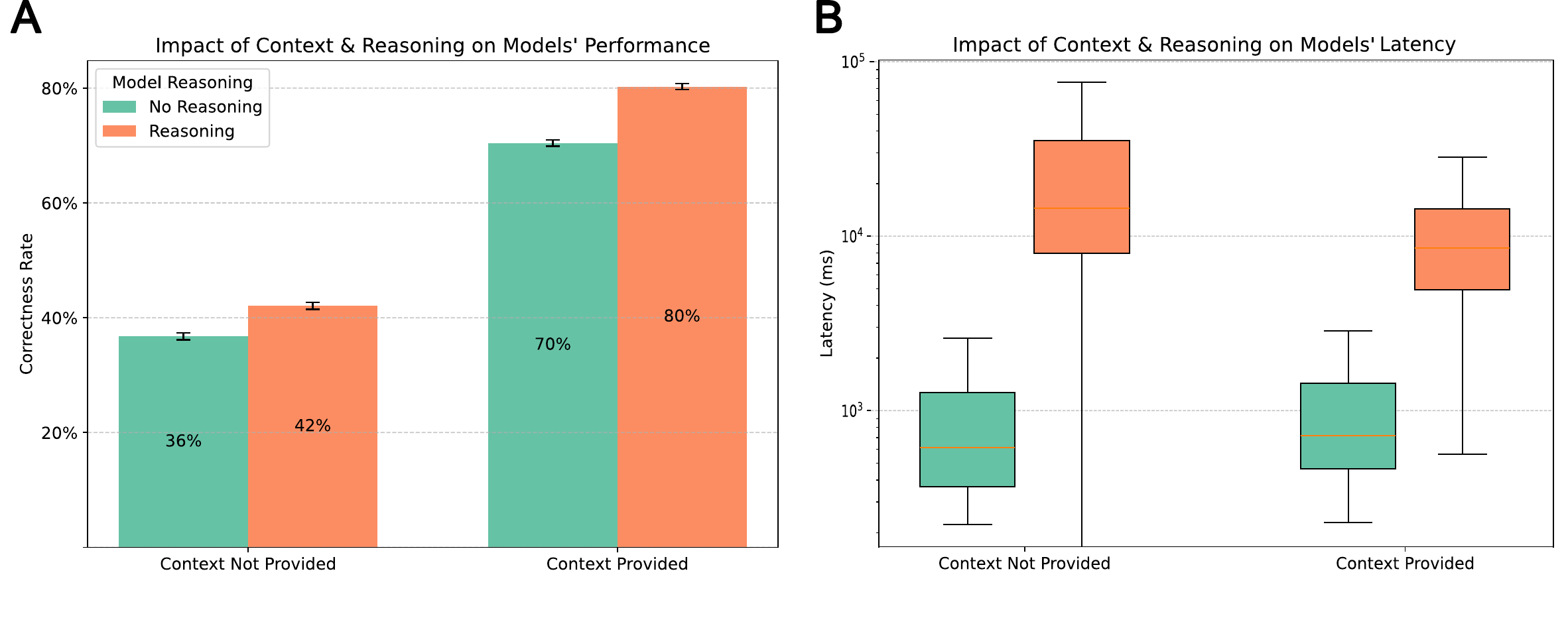}
    \caption{Impact of reasoning and context on models' Correctness Rate (left panel) and latency (right panel). Error bars represent the standard error of the mean for the models in each category.}
    \label{fig:accuracy_context_reasoning}
\end{figure}

In this analysis, we compare the performance of reasoning and non-reasoning models, i.e., models with and without test-time reasoning capabilities, respectively, in two scenarios: with context provided and without context provided. 
As illustrated in Figure~\ref{fig:accuracy_context_reasoning}-A, providing context significantly improves model performance, nearly doubling the correctness of both reasoning and non-reasoning models. Additionally, reasoning models outperform non-reasoning models in correctly answering questions, benefiting further from their reasoning capabilities in the \textit{context-provided} setup.
As expected and shown in Figure~\ref{fig:accuracy_context_reasoning}B, non-reasoning models are faster than reasoning models.
Although these models benefit significantly from context to improve their performance, their latency did not significantly change when context was provided to the model. This could be explained by the fact that the number of output tokens in these models is not sensitive to the length of input prompt and availability of context (see Figure~\ref{fig:token_usage_context_reasoning}).

For reasoning models, we observed that the availability of context lowers latency and output token count, likely because available context streamlines the thought process. For further analysis of context and reasoning in these models, refer to Appendix~\ref{sec:appendix}.

\subsection{Impact of the Number of Hops}
\begin{figure}[hb]
    \centering
    \includegraphics[width=.99\linewidth]{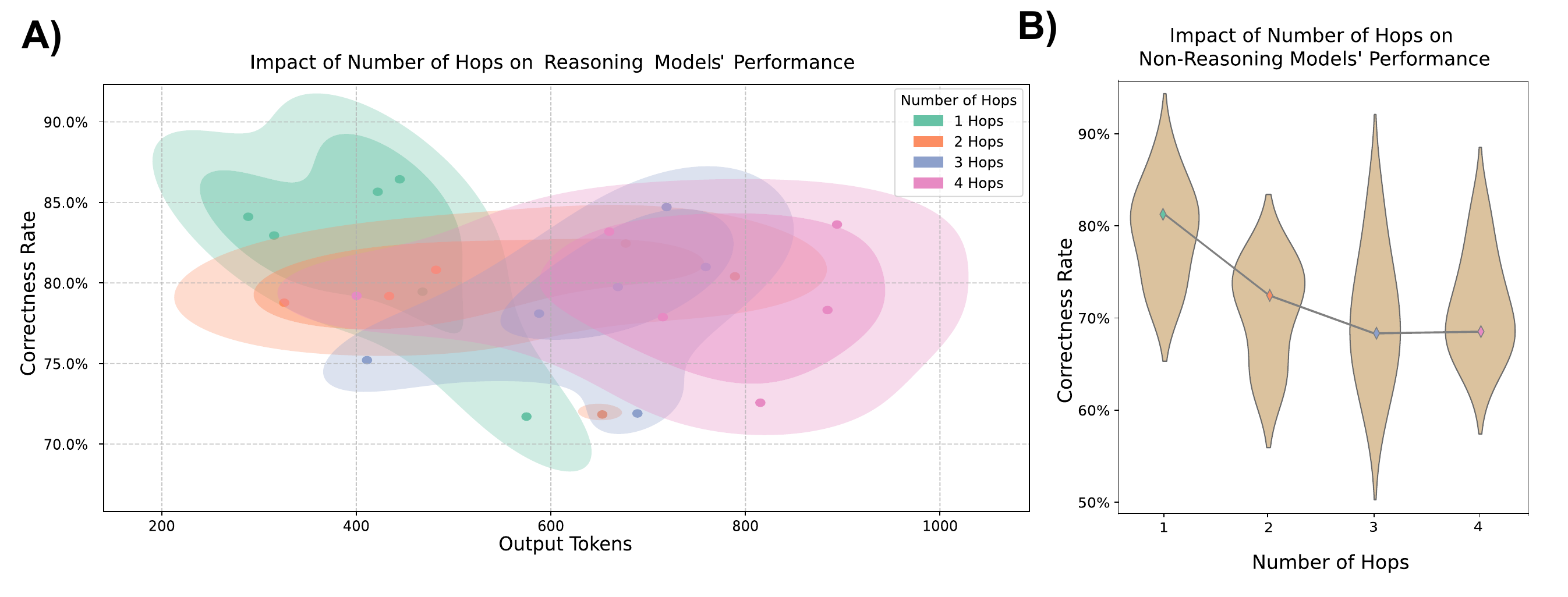}
    \caption{Analysis of the impact of the number of hops on models' performance in \textit{Context Provided} setup. panel \textit{A} illustrates the relationship between output token usage (x-axis) and correctness rate (y-axis) for varying numbers of hops. Each point represents a model, while the colored areas indicate distribution estimates for each hop count. Panel B, depicts the distribution of correctness rates for non-reasoning models across questions with different numbers of hops, with the dots representing the median of each distribution. }
    \label{fig:hops}
\end{figure}

In this section, we investigate how the number of reasoning hops influences the correctness rate and the output token count. Figure~\ref{fig:hops} presents the results for the setup with \textit{Context Provided}. For reasoning models, Figure~\ref{fig:hops}-A illustrates the distribution of answer correctness in relation to the number of generated output tokens. 
The first observation is that as the number of hops increases, the output token count, which reflects the number of thinking tokens, also increases. However, the answer correctness rate remains relatively constant for multi-hop questions, albeit slightly lower than that observed in single-hop scenarios. For single-hop questions with context provided, we see a decrease in the correctness rate as the output token count increases, indicating a performance trade-off associated with deeper reasoning in simple questions. These trends are not present when context is not provided to the models, as shown in Supplementary Figure~\ref{fig:hops_no_context}-A).



In non-reasoning models, the output token count is not sensitive to the context or complexity of the question; thus, these models are evaluated solely based on answer correctness. Figure \ref{fig:hops}-B depicts the distribution of answer correctness rate across the evaluated reasoning models for different numbers of hops. Single-hop questions are answered at a higher correctness rate than multi-hop questions, a trend not observed as clearly in the context-free setup (Supplementary figure \ref{fig:hops_no_context}-B).

\section{Conclusion}

In this study, we developed a domain-specific multi-hop question-answering (QA) system and evaluated state-of-the-art large language models within the chemistry domain.
 Our findings reveal that these models struggle with in-domain multi-hop scientific questions, correctly answering fewer than half of the queries when context is unavailable. Although reasoning fine-tuned models show marginally improved performance, they still face significant challenges. The incorporation of context leads to substantial enhancements, nearly doubling the performance of both reasoning and non-reasoning models. However, even with context, no model, including those fine-tuned for reasoning, achieved a perfect score.
Additionally, we contribute to the field by proposing an automated pipeline that integrates advanced named entity recognition with knowledge graph construction to generate intricate multi-hop reasoning tasks, which were utilized for the benchmark. Notably, this potentially domain-agnostic framework can be adapted for various fields by adjusting the pipeline to suit the target domain -- such as replacing chemistry-specific named entity recognition with appropriate alternatives-- thereby laying a robust foundation for future research aimed at enhancing reasoning capabilities across diverse specialized domains.

\paragraph{Limitations} 

Like all research, our study has certain limitations. Our benchmarking was conducted in two setups: without context and with full context provided. However, real-world applications typically involve step-by-step context collection throughout the reasoning process, leading to the possibility of partial context retrieval. Non-reasoning models often rely on a single retrieval round prior to generating answers, which does not reflect this iterative nature. Implementing a retrieval-augmented generation (RAG) pipeline with multi-step retrieval could better align with practical scenarios, bridging the gap between oracle and no-context performance. Recent advancements in reasoning models have explored integrating generative models with RAG systems for multi-step retrieval \citep{tang2024multihop, shi2024generate, zhang2024hierarchical, liu2025hoprag}. Future work should focus on developing and validating an industrial-grade, multi-step retrieval pipeline specifically for chemical texts, leveraging incremental context acquisition and structured reasoning to enhance accuracy in addressing complex chemical queries.

\paragraph{Acknowledgements}
The author(s) gratefully acknowledge the financial support for the research, authorship, and/or publication of this article provided by \textbf{MITACS} under funding number IT32409. We also thank \textbf{Adam Wojciech Bartwiki} for his leadership and project management expertise; \textbf{Tobias Roth} for his essential contributions to infrastructure; and \textbf{Stephen Dokas} for his invaluable recommendations in chemistry.

\bibliographystyle{unsrt}
\bibliography{bib}

\begin{thebibliography}{10}

\bibitem{wei2022chain}
Jason Wei, Xuezhi Wang, Dale Schuurmans, Maarten Bosma, Fei Xia, Ed~Chi, Quoc~V Le, Denny Zhou, et~al.
\newblock Chain-of-thought prompting elicits reasoning in large language models.
\newblock {\em Advances in neural information processing systems}, 35:24824--24837, 2022.

\bibitem{wang2022self}
Xuezhi Wang, Jason Wei, Dale Schuurmans, Quoc Le, Ed~Chi, Sharan Narang, Aakanksha Chowdhery, and Denny Zhou.
\newblock Self-consistency improves chain of thought reasoning in language models.
\newblock {\em arXiv preprint arXiv:2203.11171}, 2022.

\bibitem{yao2023tree}
Shunyu Yao, Dian Yu, Jeffrey Zhao, Izhak Shafran, Tom Griffiths, Yuan Cao, and Karthik Narasimhan.
\newblock Tree of thoughts: Deliberate problem solving with large language models.
\newblock {\em Advances in neural information processing systems}, 36:11809--11822, 2023.

\bibitem{besta2024graph}
Maciej Besta, Nils Blach, Ales Kubicek, Robert Gerstenberger, Michal Podstawski, Lukas Gianinazzi, Joanna Gajda, Tomasz Lehmann, Hubert Niewiadomski, Piotr Nyczyk, et~al.
\newblock Graph of thoughts: Solving elaborate problems with large language models.
\newblock In {\em Proceedings of the AAAI Conference on Artificial Intelligence}, volume~38, pages 17682--17690, 2024.

\bibitem{xiang2025towards}
Violet Xiang, Charlie Snell, Kanishk Gandhi, Alon Albalak, Anikait Singh, Chase Blagden, Duy Phung, Rafael Rafailov, Nathan Lile, Dakota Mahan, et~al.
\newblock Towards system 2 reasoning in llms: Learning how to think with meta chain-of-though.
\newblock {\em arXiv preprint arXiv:2501.04682}, 2025.

\bibitem{lewis2020retrieval}
Patrick Lewis, Ethan Perez, Aleksandra Piktus, Fabio Petroni, Vladimir Karpukhin, Naman Goyal, Heinrich K{\"u}ttler, Mike Lewis, Wen-tau Yih, Tim Rockt{\"a}schel, et~al.
\newblock Retrieval-augmented generation for knowledge-intensive nlp tasks.
\newblock {\em Advances in neural information processing systems}, 33:9459--9474, 2020.

\bibitem{d2020neurosymbolic}
Artur d'Avila Garcez and Luis~C Lamb.
\newblock Neurosymbolic ai: the 3rd wave.
\newblock {\em arXiv e-prints}, pages arXiv--2012, 2020.

\bibitem{santoro2017simple}
Adam Santoro, David Raposo, David~G Barrett, Mateusz Malinowski, Razvan Pascanu, Peter Battaglia, and Timothy Lillicrap.
\newblock A simple neural network module for relational reasoning.
\newblock {\em Advances in neural information processing systems}, 30, 2017.

\bibitem{openaiO1SystemCard}
OpenAI.
\newblock Openai o1 system card, 2024.
\newblock Accessed: 2025-03-20.

\bibitem{openaiO3MiniSystemCard}
OpenAI.
\newblock Openai o3 mini system card, 2024.
\newblock Accessed: 2025-03-20.

\bibitem{zelikman2022star}
Eric Zelikman, Yuhuai Wu, Jesse Mu, and Noah Goodman.
\newblock Star: Bootstrapping reasoning with reasoning.
\newblock {\em Advances in Neural Information Processing Systems}, 35:15476--15488, 2022.

\bibitem{guo2025deepseek}
Daya Guo, Dejian Yang, Haowei Zhang, Junxiao Song, Ruoyu Zhang, Runxin Xu, Qihao Zhu, Shirong Ma, Peiyi Wang, Xiao Bi, et~al.
\newblock Deepseek-r1: Incentivizing reasoning capability in llms via reinforcement learning.
\newblock {\em arXiv preprint arXiv:2501.12948}, 2025.

\bibitem{patil2025advancing}
Avinash Patil.
\newblock Advancing reasoning in large language models: Promising methods and approaches.
\newblock {\em arXiv preprint arXiv:2502.03671}, 2025.

\bibitem{cobbe2021training}
Karl Cobbe, Vineet Kosaraju, Mohammad Bavarian, Mark Chen, Heewoo Jun, Lukasz Kaiser, Matthias Plappert, Jerry Tworek, Jacob Hilton, Reiichiro Nakano, et~al.
\newblock Training verifiers to solve math word problems.
\newblock {\em arXiv preprint arXiv:2110.14168}, 2021.

\bibitem{hendrycks2021measuring}
Dan Hendrycks, Collin Burns, Saurav Kadavath, Akul Arora, Steven Basart, Eric Tang, Dawn Song, and Jacob Steinhardt.
\newblock Measuring mathematical problem solving with the math dataset.
\newblock {\em arXiv preprint arXiv:2103.03874}, 2021.

\bibitem{chen2021evaluating}
Mark Chen, Jerry Tworek, Heewoo Jun, Qiming Yuan, Henrique Ponde De~Oliveira Pinto, Jared Kaplan, Harri Edwards, Yuri Burda, Nicholas Joseph, Greg Brockman, et~al.
\newblock Evaluating large language models trained on code.
\newblock {\em arXiv preprint arXiv:2107.03374}, 2021.

\bibitem{austin2021program}
Jacob Austin, Augustus Odena, Maxwell Nye, Maarten Bosma, Henryk Michalewski, David Dohan, Ellen Jiang, Carrie Cai, Michael Terry, Quoc Le, et~al.
\newblock Program synthesis with large language models.
\newblock {\em arXiv preprint arXiv:2108.07732}, 2021.

\bibitem{jimenez2023swe}
Carlos~E Jimenez, John Yang, Alexander Wettig, Shunyu Yao, Kexin Pei, Ofir Press, and Karthik Narasimhan.
\newblock Swe-bench: Can language models resolve real-world github issues?
\newblock {\em arXiv preprint arXiv:2310.06770}, 2023.

\bibitem{yang2018hotpotqa}
Zhilin Yang, Peng Qi, Saizheng Zhang, Yoshua Bengio, William~W Cohen, Ruslan Salakhutdinov, and Christopher~D Manning.
\newblock Hotpotqa: A dataset for diverse, explainable multi-hop question answering.
\newblock {\em arXiv preprint arXiv:1809.09600}, 2018.

\bibitem{geva2021did}
Mor Geva, Daniel Khashabi, Elad Segal, Tushar Khot, Dan Roth, and Jonathan Berant.
\newblock Did aristotle use a laptop? a question answering benchmark with implicit reasoning strategies.
\newblock {\em Transactions of the Association for Computational Linguistics}, 9:346--361, 2021.

\bibitem{wellawatte2024chemlit}
Geemi Wellawatte, Huixuan Guo, Magdalena Lederbauer, Anna Borisova, Matthew Hart, Marta Brucka, and Philippe Schwaller.
\newblock Chemlit-qa: A human evaluated dataset for chemistry rag tasks.
\newblock In {\em AI for Accelerated Materials Design-NeurIPS 2024}.

\bibitem{huang2024olympicarena}
Zhen Huang, Zengzhi Wang, Shijie Xia, and Pengfei Liu.
\newblock Olympicarena medal ranks: Who is the most intelligent ai so far?
\newblock {\em arXiv preprint arXiv:2406.16772}, 2024.

\bibitem{rein2024gpqa}
David Rein, Betty~Li Hou, Asa~Cooper Stickland, Jackson Petty, Richard~Yuanzhe Pang, Julien Dirani, Julian Michael, and Samuel~R Bowman.
\newblock Gpqa: A graduate-level google-proof q\&a benchmark.
\newblock In {\em First Conference on Language Modeling}, 2024.

\bibitem{zheng2025large}
Zhiling Zheng, Nakul Rampal, Theo~Jaffrelot Inizan, Christian Borgs, Jennifer~T Chayes, and Omar~M Yaghi.
\newblock Large language models for reticular chemistry.
\newblock {\em Nature Reviews Materials}, pages 1--13, 2025.

\bibitem{mavi2024multi}
Vaibhav Mavi, Anubhav Jangra, Adam Jatowt, et~al.
\newblock Multi-hop question answering.
\newblock {\em Foundations and Trends{\textregistered} in Information Retrieval}, 17(5):457--586, 2024.

\bibitem{welbl2018constructing}
Johannes Welbl, Pontus Stenetorp, and Sebastian Riedel.
\newblock Constructing datasets for multi-hop reading comprehension across documents.
\newblock {\em Transactions of the Association for Computational Linguistics}, 6:287--302, 2018.

\bibitem{trivedi2022musique}
Harsh Trivedi, Niranjan Balasubramanian, Tushar Khot, and Ashish Sabharwal.
\newblock Musique: Multihop questions via single-hop question composition.
\newblock {\em Transactions of the Association for Computational Linguistics}, 10:539--554, 2022.

\bibitem{tang2024multihop}
Yixuan Tang and Yi~Yang.
\newblock Multihop-rag: Benchmarking retrieval-augmented generation for multi-hop queries.
\newblock {\em arXiv preprint arXiv:2401.15391}, 2024.

\bibitem{mirza2025framework}
Adrian Mirza, Nawaf Alampara, Sreekanth Kunchapu, Marti{\~n}o R{\'\i}os-Garc{\'\i}a, Benedict Emoekabu, Aswanth Krishnan, Tanya Gupta, Mara Schilling-Wilhelmi, Macjonathan Okereke, Anagha Aneesh, et~al.
\newblock A framework for evaluating the chemical knowledge and reasoning abilities of large language models against the expertise of chemists.
\newblock {\em Nature Chemistry}, pages 1--8, 2025.

\bibitem{melnyk2022knowledge}
Igor Melnyk, Pierre Dognin, and Payel Das.
\newblock Knowledge graph generation from text.
\newblock {\em arXiv preprint arXiv:2211.10511}, 2022.

\bibitem{zhang2024extract}
Bowen Zhang and Harold Soh.
\newblock Extract, define, canonicalize: An llm-based framework for knowledge graph construction.
\newblock {\em arXiv preprint arXiv:2404.03868}, 2024.

\bibitem{das2018building}
Rajarshi Das, Tsendsuren Munkhdalai, Xingdi Yuan, Adam Trischler, and Andrew McCallum.
\newblock Building dynamic knowledge graphs from text using machine reading comprehension.
\newblock {\em arXiv preprint arXiv:1810.05682}, 2018.

\bibitem{langer2024cear}
Stefan Langer, Fabian Neuhaus, and Andreas N{\"u}rnberger.
\newblock Cear: Automatic construction of a knowledge graph of chemical entities and roles from scientific literature.
\newblock {\em arXiv preprint arXiv:2407.21708}, 2024.

\bibitem{liao2023coarse}
Wenxiong Liao, Zhengliang Liu, Yiyang Zhang, Xiaoke Huang, Fei Qi, Siqi Ding, Hui Ren, Zihao Wu, Haixing Dai, Sheng Li, et~al.
\newblock Coarse-to-fine knowledge graph domain adaptation based on distantly-supervised iterative training.
\newblock In {\em 2023 IEEE International Conference on Bioinformatics and Biomedicine (BIBM)}, pages 1294--1299. IEEE, 2023.

\bibitem{ruas2022nilinker}
Pedro Ruas and Francisco~M Couto.
\newblock Nilinker: attention-based approach to nil entity linking.
\newblock {\em Journal of Biomedical Informatics}, 132:104137, 2022.

\bibitem{pubmedbert}
Yu~Gu, Robert Tinn, Hao Cheng, Michael Lucas, Naoto Usuyama, Xiaodong Liu, Tristan Naumann, Jianfeng Gao, and Hoifung Poon.
\newblock Domain-specific language model pretraining for biomedical natural language processing, 2020.

\bibitem{kim2021pubchem}
Sunghwan Kim, Jie Chen, Tiejun Cheng, Asta Gindulyte, Jia He, Siqian He, Qingliang Li, Benjamin~A Shoemaker, Paul~A Thiessen, Bo~Yu, et~al.
\newblock Pubchem in 2021: new data content and improved web interfaces.
\newblock {\em Nucleic acids research}, 49(D1):D1388--D1395, 2021.

\bibitem{shi2024generate}
Zhengliang Shi, Weiwei Sun, Shen Gao, Pengjie Ren, Zhumin Chen, and Zhaochun Ren.
\newblock Generate-then-ground in retrieval-augmented generation for multi-hop question answering.
\newblock {\em arXiv preprint arXiv:2406.14891}, 2024.

\bibitem{zhang2024hierarchical}
Xiaoming Zhang, Ming Wang, Xiaocui Yang, Daling Wang, Shi Feng, and Yifei Zhang.
\newblock Hierarchical retrieval-augmented generation model with rethink for multi-hop question answering.
\newblock {\em arXiv preprint arXiv:2408.11875}, 2024.

\bibitem{liu2025hoprag}
Hao Liu, Zhengren Wang, Xi~Chen, Zhiyu Li, Feiyu Xiong, Qinhan Yu, and Wentao Zhang.
\newblock Hoprag: Multi-hop reasoning for logic-aware retrieval-augmented generation.
\newblock {\em arXiv preprint arXiv:2502.12442}, 2025.

\end{thebibliography}

\newpage
\section{Appendix}

\label{sec:appendix}
\makeatletter
\renewcommand \thesection{S\@arabic\c@section}
\renewcommand\thetable{S\@arabic\c@table}
\renewcommand \thefigure{S\@arabic\c@figure}
\makeatother

\newcommand{\hlblue}[1]{\colorbox{blue!15}{#1}}
\newcommand{\hlgreen}[1]{\colorbox{green!15}{#1}}
\newcommand{\hlred}[1]{\colorbox{red!15}{#1}}

\subsection{Detailed Knowledge Graph Generation}
In this section, we explain each step in our graph generation process, illustrating how unstructured chemical text is transformed into a structured representation suitable for downstream tasks.

\subsubsection{Text Preprocessing}
Our preprocessing pipeline is designed to isolate the most informative portions of ChemRxiv articles and prepare them for entity and relation extraction. First, we retrieved all ChemRxiv articles whose licenses permit redistribution via the ChemRxiv API. Using regular expressions, we extracted the introduction section of each paper, as this typically focuses on concise, objective statements about chemical phenomena. From each introduction, we extracted the first 500 words; this limit balances coverage of key background information against the risk of including less relevant or overly general text. To avoid fragmenting paragraphs, we ensured that no paragraph was split across chunks. Finally, we segmented each introduction into contiguous chunks of up to 128 words, preserving natural sentence boundaries. This segmentation reduces the input length for subsequent processing steps, ensuring that each chunk remains within the token limits of our downstream models while retaining semantic coherence.

\subsubsection{Node Extraction}
Once the text is segmented, we identify chemical entities and their interrelations. We apply a named entity recognition (NER) model based on the PubMedBERT architecture, fine-tuned on multiple chemical datasets, to detect candidate entities within each 128-word chunk. To further improve precision and eliminate spurious mentions, we refine the NER output using OpenAI’s \texttt{gpt-4o}, prompting it with a few shots to verify whether each detected span indeed corresponds to a chemically meaningful entity and to standardize entity labels (for instance, converting "MeOH" to "methanol"). Below, we showed the prompt for Entity verification.

\begin{framed}
You are a chemistry expert specializing in entity recognition. Your task is to \textbf{validate and filter} the extracted entities, ensuring they are \textbf{chemically meaningful} based on the provided text. Remove any irrelevant terms, including general descriptors, numerical values, reaction conditions, and vague terms.

\vspace{1em}

\noindent
\textbf{Entities Extracted by NER:} \\
\{\texttt{entities}\}

\vspace{1em}

\noindent
\textbf{Text for Context:} \\
\{\texttt{text}\}

\vspace{1em}

\noindent
\textbf{Criteria for Valid Entities:}
\begin{itemize}[leftmargin=1.5em]
    \item[$\checkmark$] Chemical compounds (e.g., "HCl", "Sodium hydroxide", "Ethanol", "Benzene")
    \item[$\checkmark$] Chemical elements (e.g., "Carbon", "Oxygen", "Cesium")
    \item[$\checkmark$] Specific catalysts, solvents, reagents (e.g., "Cs\textsubscript{2}CO\textsubscript{3}", "Toluene", "Palladium")
\end{itemize}

\vspace{0.5em}

\noindent
\textbf{Remove the Following Types of Entities:}
\begin{itemize}[leftmargin=1.5em]
    \item[$\times$] Generic terms (e.g., "Reaction", "Solvent", "Acid", "Base", "Solution")
    \item[$\times$] Experimental conditions (e.g., "pH", "Temperature", "2~M", "Strong acid")
    \item[$\times$] Measurement terms (e.g., "X-ray diffraction", "NMR")
    \item[$\times$] General descriptors (e.g., "High concentration", "Low efficiency")
\end{itemize}

\vspace{1em}

\noindent
\textbf{Output Format:} \\
Return only a \textbf{Python list} of valid chemical entities, with no explanations, markdown, or extra formatting.
\end{framed}

\subsubsection{Edge Extraction}

After obtaining a vetted set of entities, we extract pairwise relations using the same \texttt{gpt-4o} instance. For each pair of entities co-occurring within a chunk, we prompt the model with a few shots to classify or generate the nature of their relationship, producing triplets of the form (entity\_A, relation, entity\_B). The result of this step is a set of entity nodes (chemical names) and directed edges (relation labels) extracted directly from the text, forming a raw triplet collection that reflects the functional associations present in the chemical literature. Below, we show the prompt for Edge Extraction.

\begin{framed}
You are an expert in chemical text analysis. Your task is to extract \textbf{only chemically meaningful relationships} between a given set of entities from the provided text.

\vspace{1em}

\noindent
\textbf{Guidelines for Relation Extraction:}
\begin{enumerate}[leftmargin=1.5em]
    \item \textbf{Entity Matching:} Consider only the entities provided in the given set. If an entity appears in the text but has no meaningful chemical relationship with another entity in the set, ignore it.
    \item \textbf{Chemically Significant Relations Only:} Extract relations that describe actual \textbf{chemical interactions, transformations, or properties} (e.g., "reacts with," "catalyzes," "dissolves in," "produces").
    \item \textbf{Factual Relations:} Only extract factual relations. Avoid observations, opinions, and findings.
    \item \textbf{Tuple Format:} Output extracted facts in the form of \textbf{(entity1, relation, entity2)}.
    \item \textbf{Avoid Generic Relations:} Exclude weak relations like "is," "are," "exists," "relates to." Focus on \textbf{specific interactions}.
\end{enumerate}

\vspace{1em}

\noindent
\textbf{Valid Relation Types (Examples):}
\begin{itemize}[leftmargin=1.5em]
    \item[$\checkmark$] "reacts with"
    \item[$\checkmark$] "catalyzes"
    \item[$\checkmark$] "binds to"
    \item[$\checkmark$] "dissolves in"
    \item[$\checkmark$] "oxidizes"
    \item[$\checkmark$] "inhibits"
    \item[$\checkmark$] "precipitates with"
    \item[$\checkmark$] "acts as a solvent for"
    \item[$\checkmark$] "is synthesized from"
\end{itemize}

\vspace{0.5em}

\noindent
\textbf{Avoid These Weak Relations:}  
Exclude relations such as "is," "are," "has," "exists."

\vspace{1em}

\noindent
\textbf{Entities Provided:} \\
\{\texttt{entities}\}

\vspace{1em}

\noindent
\textbf{Text:} \\
\{\texttt{text}\}

\vspace{1em}

\noindent
\textbf{Extract at most \{\texttt{max\_facts}\} factual statements.}

\vspace{1em}

\noindent
\textbf{Output Format:} \\
Provide the output as a \textbf{Python list of tuples}, containing only the extracted relationships without any code formatting, backticks, or markdown.

\vspace{1em}

\noindent
\textbf{Example Output:}
\begin{itemize}[leftmargin=1.5em]
    \item[–] [  
      ("HCl", "dissolves in", "Water"),  
      ("HCl", "reacts with", "Sodium hydroxide")  
      ]
\end{itemize}
\end{framed}
\subsubsection{Knowledge Enrichment}
To enrich each node with descriptive metadata and resolve naming ambiguities, we retrieved supplemental information from two external resources: Wikipedia and PubChem. From Wikipedia, we extracted the introductory summary of each entity’s page, providing a concise description of its common usage, historical context, or primary function. From PubChem, we obtained the official compound name (Record Title) alongside additional names and identifiers drawn from the "Names and Identifiers" section. We retrieved a textual description from the "Record Description" heading, summarizing key properties or applications. Safety annotations, including hazard statements or pictograms, were collected from the "Chemical Safety" subsection. Structural information was captured in the form of the canonical SMILES string under "Computed Descriptors," and the molecular formula was fetched from the "Molecular Formula" field. Finally, we extracted computed physicochemical properties, such as molecular weight, topological polar surface area, and log P values, from the "Computed Properties" list within "Chemical and Physical Properties." All of these metadata fields were stored as additional text and added as external information for each of the nodes.

\subsubsection{Graph Generation}
At this stage, we constructed the graph by forming triplets of the form (node, edge, node). Each node was linked to the original text segment from which it was extracted and, if available, to its corresponding external metadata. Likewise, each edge was associated with the specific text chunk that produced it. In this way, both nodes and edges maintain direct references to their source text, ensuring traceability throughout the knowledge graph.

\subsection{Detailed Question Generation}

\subsubsection{Path Sampling from the Knowledge Graph}
To generate multi-hop questions, we first sampled paths of varying lengths from the constructed knowledge graph using a randomized breadth-first search (BFS) path sampling algorithm. During path sampling, we enforced that each edge in a sampled path originates from a distinct source text, thereby encouraging the integration of information from multiple, separate context passages. Concretely, a path of length \(K\) comprises \(K+1\) entities and traces through \(K\) different source texts extracted from the original ChemRxiv database. By requiring unique sources for each edge, we ensure that correct solutions must draw on evidence scattered across several documents rather than a single paragraph.

\subsubsection{One-Hop Question Formulation}
Adopting a bottom-up approach, we generated individual one-hop questions for each edge along a sampled path. For every triplet \((\text{entity}_1, \text{relation}, \text{entity}_2)\), we framed a question whose answer is \(\text{entity}_1\) by asking, "Which entity holds the \(\text{relation}\) relation to \(\text{entity}_2\)?" If the initial phrasing lacked sufficient specificity or clarity, we invoked a language model (OpenAI’s \texttt{o3-mini}) to augment the prompt with metadata drawn from the original text segment, such as contextual phrases or qualifying details. This enrichment step ensures that each one-hop question remains clear, precise, and answerable from the associated source text alone. Below is the prompt for generating the one-hop questions.

\begin{framed}
You are given a text along with an entity and its relation to another entity.\\

Entity 1: \{\texttt{entity1}\} \\
Relation: \{\texttt{relation}\} \\
Entity 2: \{\texttt{entity2}\} \\
Text: \{\texttt{text}\} \\

Information about Entity1: \{\texttt{entity1\_meta} if \texttt{entity1\_meta} else \texttt{None}\} \\

Your task is to generate a factual question whose answer is Entity1. \\
The question should ask for the entity that has the specified relation to Entity2. \\
Do not mention the answer (which is Entity1) in the question. \\
Ensure that the question is factual and can be answered solely based on the given text and the information about Entity1. \\
Do not refer to sections such as "Abstract," "Table \#1," "in the text," or "in the article." \\

If Entity1 and relation are not specific enough (i.e., multiple answers are possible), add descriptions from the text or from the information about Entity1 to make it specific so that Entity1 is the only answer. \\

Return a dictionary without any code formatting, backticks, or markdown, with keys \texttt{"q"} and \texttt{"a"}.
\end{framed}

\subsubsection{Multi-Hop Question Aggregation}
Once all one-hop questions for a path were vetted, we combined them into a single multi-hop question by prompting OpenAI’s \texttt{o3-mini} model, with a few shots. The final aggregated question is constructed by starting with the sub-question corresponding to the last edge (i.e., the entity nearest to the "tail" of the path) and then chaining backward through each preceding entity until reaching \(\text{entity}_1\) of the first relation. In other words, if the sampled path is 
\[
(\text{entity}_1 \xrightarrow{\text{relation}_1} \text{entity}_2),\;(\text{entity}_2 \xrightarrow{\text{relation}_2} \text{entity}_3),\;\dots,\;(\text{entity}_K \xrightarrow{\text{relation}_K} \text{entity}_{K+1}),
\]
the aggregated question asks first about \(\text{entity}_{K+1}\) (the final tail), then uses its answer as context for the penultimate question, and so on, so that the final answer corresponds to \(\text{entity}_1\). This reverse-chaining structure ensures that the multi-hop prompt leads directly to the original target node while preserving logical flow. Below, we show the prompt for generating the multi-hop question.

\begin{framed}
You are given multiple factual questions and their answers that are logically connected.\\
Your task is to chain them into a single, coherent multi-hop question that requires multiple reasoning steps.\\
Ensure that the (only) answer is the answer to the first question, and the question naturally follows from the facts given.\\
You have to start from the last generated question and build up a single multi-hop question so it aggregates them all and the answer is the answer to the first question.\\
None of the answers to any of the questions should be in the generated question.\\

Here is an example: \\
Example: \\
Q1: What is oxidized to form Carbon Dioxide? \\
A1: Methane \\
Q2: What is used in Photosynthesis? \\
A2: Carbon Dioxide \\
Q3: What produces Oxygen? \\
A3: Photosynthesis \\

Multi-hop question: \\
Q: What is oxidized to produce a substance that is used in a process that results in Oxygen? \\
A: Methane \\

Here are the generated questions and answers: \\
\{\texttt{formatted\_qas}\} \\

Return a Python dictionary without any code formatting, backticks, or markdown, with keys \texttt{"q"} (multi-hop question) and \texttt{"a"} (final answer).
\end{framed}

\subsubsection{Verification and Filtering}
During verification, we first reviewed each one-hop question for clarity, chemical relevance, and direct answerability based on its corresponding source text. Below is the prompt for one-hop question evaluation.

\begin{framed}
You are a chemistry expert. Your task is to determine if the given question is a factual chemistry question, unambiguous (has only one answer), and answerable based on the provided context. A factual question must be based on actual chemical properties, reactions, or experimentally verified principles and must be strictly related to chemistry. An answerable question should be solvable based on the given context and must not be open-ended or have multiple correct answers. MAKE SURE THE QUESTION HAS ONLY ONE CORRECT ANSWER. There shouldn’t be any other entity except for the given answer that could be another answer. \\

\#\#\# Question: \\
\{\texttt{question}\} \\

\#\#\# Answer: \\
\{\texttt{answer}\} \\

\#\#\# Context: \\
\{\texttt{context}\} \\

Please analyze the context and verify if the question is factual, unambiguous, and answerable. If the question is factual, has only one correct answer, is strictly related to chemistry, and can be answered based on the context, return "yes." Otherwise, return "no." \\

\#\#\# Examples of Factual Chemistry Questions: \\
\cmark "What dissolves in water and evaporates at 0 °C?" \\
\cmark "What catalyst is used in the reaction between A and B?" \\

\#\#\# Examples of Non-Factual or Ambiguous Chemistry Questions: \\
\xmark\xmark "What is the song of Nirvana that is a chemical entity?" \\
\xmark\xmark "What chemical entity and structural unit form the layered hydroxide structures with intercalated water ions used in battery materials and OER catalysis?" (M(OH)\textsubscript{6} and $\alpha$-Ni(OH)\textsubscript{2} are valid answers) \\
\xmark\xmark Questions that have multiple possible correct answers or are not strictly related to chemistry.
\end{framed}

Next, the multi-hop question underwent an additional evaluation step to confirm that the logical chain, from the final sub-question back to the first, correctly guides a reader (or model) to \(\text{entity}_1\). We employed an LLM-based verification process to assess factual accuracy, answerability given available context and metadata, and overall coherence among sub-questions. Feedback from domain experts was continuously incorporated into the prompt templates to refine verification accuracy. Any question, either one-hop or multi-hop, that was answered incorrectly by all evaluated models was excluded from the benchmark to minimize ambiguity and ensure high-quality, unambiguous reasoning tasks. Below is the prompt for evaluating the whole path.

\begin{framed}
You are a chemistry expert. Your task is to determine if the given question is a factual chemistry question and answerable based on the provided path.\\

\#\#\# Path Information: \\
\{\texttt{path\_text}\} \\

\#\#\# Question: \\
\{\texttt{question}\} \\

\#\#\# Answer: \\
\{\texttt{answer}\} \\

Please analyze the path and verify if the question is a factual chemistry question and can be answered based on the given path. A factual question must be based on actual chemical properties, reactions, or experimentally verified principles. An answerable question should be solvable based on the given path. If the question is factual and answerable, return "yes". If it contains speculation, opinions, or lacks verifiable chemical grounding, or it is not solvable, return "no".\\

\#\#\# Examples of Factual Chemistry Questions: \\
\cmark "What dissolves in water?" \\
\cmark "What catalyst is used in the reaction between A and B?" \\
\cmark "Which compound undergoes oxidation in this reaction?" \\
\cmark "What product is formed when sodium reacts with chlorine?" \\

\#\#\# Examples of Non-Factual Chemistry Questions: \\
\xmark\xmark "Why do some scientists think this reaction is inefficient?" \\
\xmark\xmark "What is the best solvent for this reaction?" \\
\xmark\xmark "Is this reaction useful in industry?" \\
\xmark\xmark "Do you think this compound is a good catalyst?" \\

Provide only "yes" or "no" as your response.
\end{framed}

\subsubsection{Short-Answer Design and Rationale}
To minimize the impact of writing style and summarization on accuracy evaluation, all questions were deliberately designed to elicit short, concise answers. Answering a multi-hop question requires decomposing it into its constitutive one-hop sub-questions, retrieving each intermediate answer, and then combining them to arrive at the final answer. Even when full context is available, a correct response cannot be produced if a model fails to infer and integrate multiple pieces of knowledge. By keeping answers brief and focusing on discrete factual steps, we ensure that performance evaluation reflects a model’s ability to conduct multi-hop inference rather than its capacity to paraphrase or generate lengthy explanations.

\subsection{Rejected Questions}
\label{subsec:rejected_q}
There were a couple of reasons that we identified for rejecting of some questions from the benchmark. One was having multiple valid answers, and second was the answer was present in the question. 
Some examples are provided below.

\begin{framed}
\textbf{Q1:}

\textbf{Context:} \\
Researchers have developed an anode material based on NiCo\textsubscript{r}GO (Nickel–Cobalt–reduced Graphene Oxide). In some variants, the NiCo\textsubscript{r}GO is further decorated with palladium (Pd) nanoparticles to enhance catalytic performance.

\vspace{0.5em}
\textbf{Question:} \\
Which component in the electrode structure functions as a catalyst at the anode when incorporating decorated NiCo\textsubscript{r}GO?

\vspace{0.5em}
\textbf{Issue:} \\
The question is declined due to ambiguity; two distinct answers are technically correct based on the variant of the material:

\begin{itemize}
    \item If the material is \textbf{NiCo\textsubscript{r}GO} without Pd, the catalyst is \textbf{Nickel (Ni)}.
    \item If the material is \textbf{Pd-decorated NiCo\textsubscript{r}GO}, the catalyst is \textbf{Palladium (Pd)}.
\end{itemize}

Since the phrasing of the question does not clearly disambiguate which material variant is being used, it leads to multiple valid interpretations. Therefore, it cannot be accepted as a single-answer question.

\textbf{Q2:}

\textbf{Context:} \\
Hypervalent iodine compounds such as diaryliodonium salts are widely used as electrophilic arylation reagents. According to the source text, these salts are employed in both \textbf{transition metal-catalyzed} and \textbf{metal-free} arylation reactions. These reactions can be used to functionalize aromatic compounds, including halogen-substituted analogues like CIMPPC, by replacing hydrogen atoms.

\vspace{0.5em}
\textbf{Question:} \\
Which type of arylation reaction that employs electrophilic arylation reagents utilizes diaryliodonium salts in hypervalent iodine chemistry?

\vspace{0.5em}
\textbf{Expected Answer:} \\
\textit{Transition metal-catalyzed arylations}

\vspace{0.5em}
\textbf{Reason for Rejection:} \\
The question is declined due to the presence of \textbf{multiple valid answers}. The source explicitly states that diaryliodonium salts are used in:
\begin{itemize}
    \item \textbf{Transition metal-catalyzed arylations}, and
    \item \textbf{Metal-free arylations}.
\end{itemize}
Both are equally valid interpretations of the question. Without further constraints or clarification, the question has more than one correct answer and does not meet the single-answer requirement.
\end{framed}

To minimize ambiguity, we excluded questions that were answered incorrectly by all evaluated models from the benchmark. A subset of these questions was manually assessed, with most categorized as having multiple valid answers. Additionally, as explained in Section~\ref{sec:qa-generation}, an overall LLM verifier evaluated all questions for quality and answer inclusion. Any questions rejected by this verifier were also removed from the pool.

\subsection{Expert feedback}
\label{subsec:expert_feedback}
As summarized in Table~\ref{tab:quality_metrics}, we report the average number of models that produced a correct answer with full context versus no context for each category. We also include the mean number of reasoning hops per question (2.46–2.60). Overall, the observations indicate that question complexity, measured by the number of hops, and the accuracy of LLM answers do not reliably predict question quality.
All annotation guidelines, per-question ratings, and raw model outputs are available in our public repository.

\begin{table}[ht]
  \centering
  \begin{tabular}{lcccc}
    \toprule
    \textbf{Cat.} 
      & \textbf{Num.\ Que.\ } 
      & \textbf{Avg.\ Corr.\ (+ctx)} 
      & \textbf{Avg.\ Corr.\ (–ctx)} 
      & \textbf{Avg.\ len.} \\
    \midrule
    Good & 26 (65\%)   & 7.5  & 4.2  & 2.46 \\
    Ok   &  9 (22.5\%) & 7.55 & 3.44 & 2.55 \\
    Poor &  5 (12.5\%) & 7.8  & 4.6  & 2.60 \\
    \bottomrule
  \end{tabular}
  \vspace{1em}
  \caption{The 40 high‐confidence questions are grouped by expert‐rated quality (\textit{Good}, \textit{Ok}, \textit{Poor}). \textbf{Num.\ Que.} shows the count and percentage of questions in each category. \textbf{Avg.\ Corr.\ (+ctx)} / \textbf{(–ctx)} gives the mean number of models that answered correctly when provided with full supporting context versus no context, highlighting how contextual evidence boosts multi‐hop inference. \textbf{Avg.\ len.} denotes the average number of reasoning hops required per question.}
  \label{tab:quality_metrics}
\end{table}

\subsection{Detailed Performance Based on Context Availability}
\begin{figure*}[ht!]
    \centering
    \includegraphics[width=\linewidth]{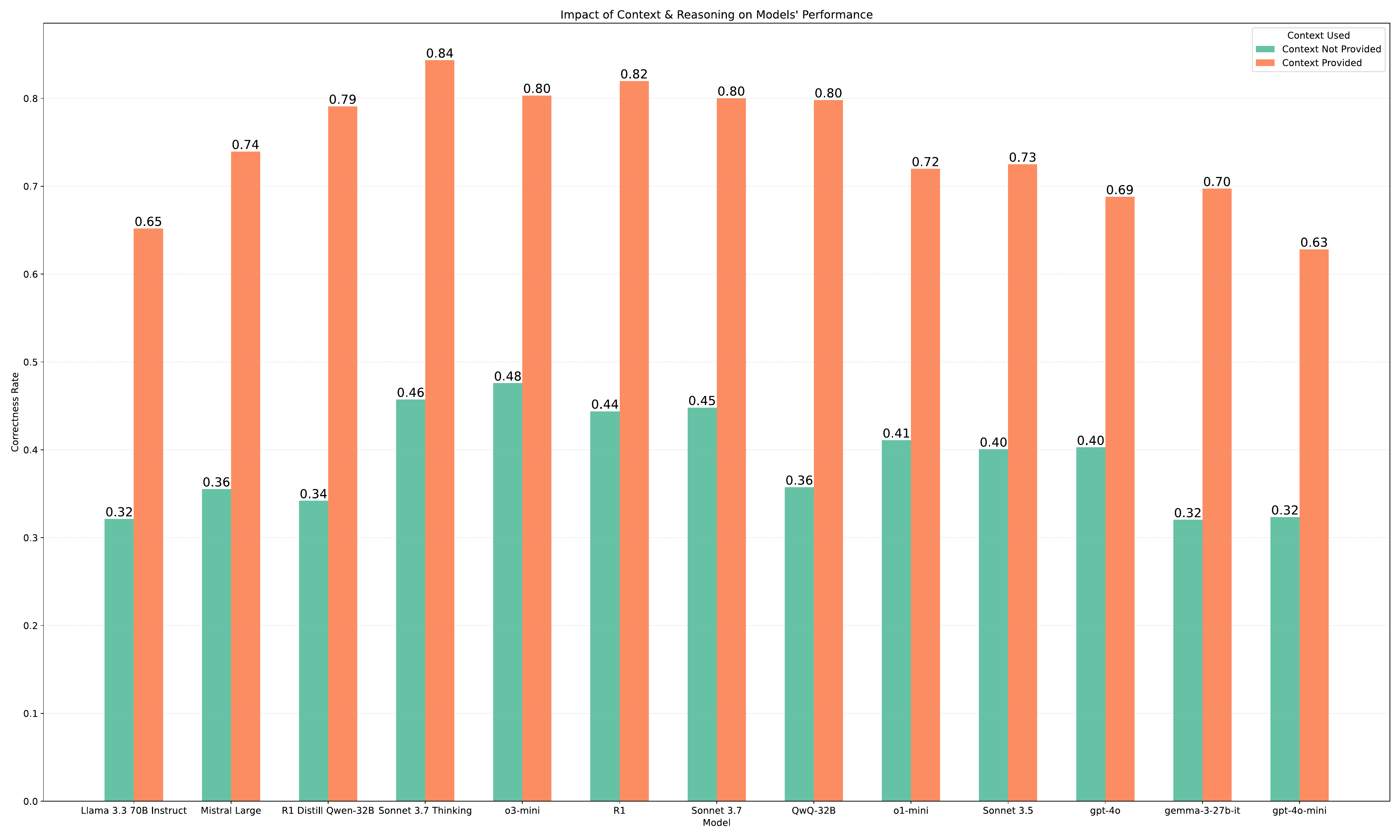}
    \caption{Correctness rate of models with respect to context.}
    \label{fig:context_accuracy}
\end{figure*}

\begin{figure*}[ht!]
    \centering
    \includegraphics[width=\linewidth]{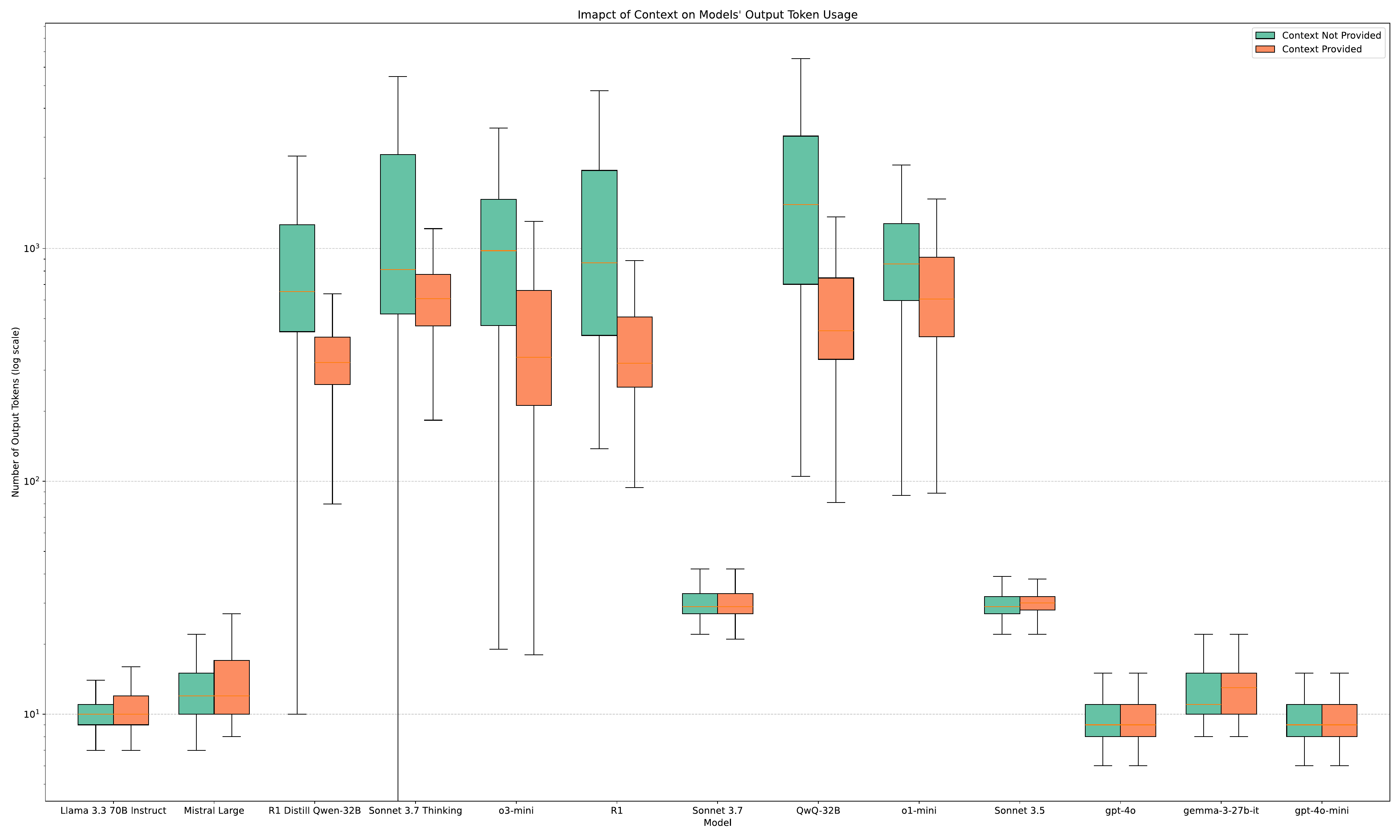}
    \caption{Token usage of models with respect to context}
    \label{fig:context_token_usage}
\end{figure*}

\begin{figure*}[ht!]
    \centering
    \includegraphics[width=\linewidth]{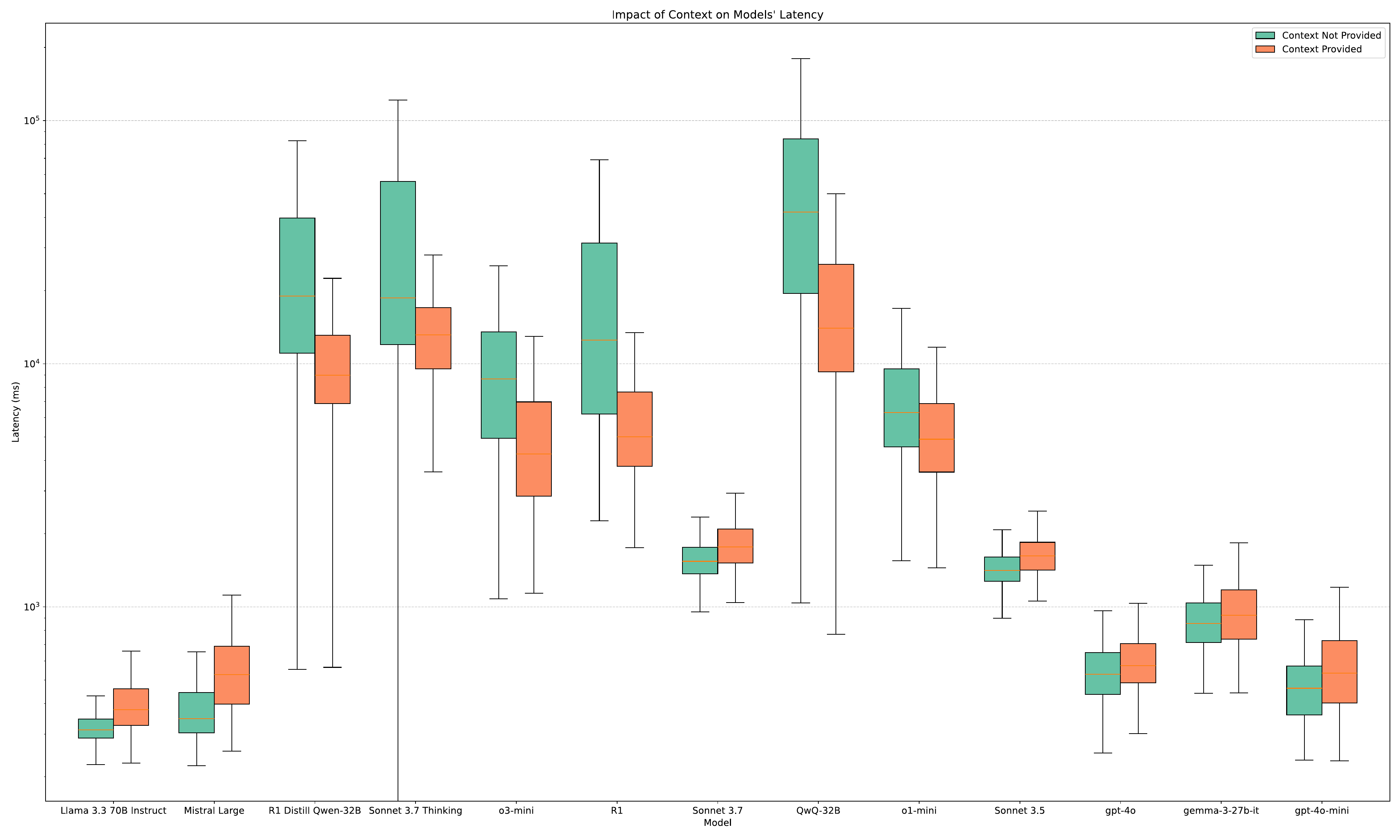}
    \caption{Latency with respect to context}
    \label{fig:context_latency}
\end{figure*}

Figure~\ref{fig:context_accuracy} shows that model performance is strongly influenced by whether context is provided in the input. In particular, Claude 3.7 with extended thinking achieved a correctness rate of 84\% with context, whereas \texttt{o3-mini} recorded the highest correctness rate (48\%) when the context was absent. Note that \texttt{o3-mini} was primarily used to generate the questions, which may have introduced a slight bias, resulting in its minor improvement in correctness. Figures~\ref{fig:context_token_usage} and~\ref{fig:context_latency} illustrate the token usage and latency of the models, respectively.

\newpage

\newpage
\newpage
\subsection{Performance of models on Chemistry Subset of HotpotQA}

Table~\ref{tab:hotpotqa_results} shows the details of each model's performance, latency, and tokens used in both setups of context provided and not provided for the chemistry subset of HotpotQA \citep{yang2018hotpotqa} questions. 

\begin{table*}[hbpt]
  \centering
  \resizebox{\textwidth}{!}{%
    \begin{tabular}{lccccccc}
      \toprule
      Model & Context & Correctness Rate (\%) & Avg Duration (s) & Avg Input Tokens & Avg Output Tokens & Total Input Tokens (K) & Total Output Tokens (K) \\
      \midrule
        Anthropic Claude Sonnet 3.5 V2 & \xmark & 53.27 & 1.26 & 517 & 29 & 507.47 & 29.20 \\
        Anthropic Claude Sonnet 3.5 V2 & \cmark & 84.80 & 1.32 & 618 & 29 & 605.73 & 28.85 \\
        Anthropic Claude Sonnet 3.7 & \xmark & 58.27 & 1.87 & 517 & 29 & 507.47 & 28.49 \\
        Anthropic Claude Sonnet 3.7 & \cmark & 86.22 & 1.94 & 618 & 29 & 605.73 & 28.71 \\
        Anthropic Claude Sonnet 3.7 (Thinking) & \xmark & 65.31 & 17.54 & 539 & 726 & 528.48 & 711.49 \\
        Anthropic Claude Sonnet 3.7 (Thinking) & \cmark & 87.14 & 10.10 & 640 & 390 & 627.29 & 382.77 \\
        OpenAI GPT-4o-mini & \xmark & 45.61 & 0.43 & 170 & 7 & 166.96 & 7.72 \\
        OpenAI GPT-4o-mini & \cmark & 80.31 & 0.50 & 257 & 8 & 252.39 & 8.00 \\
        OpenAI GPT-4o & \xmark & 55.10 & 0.62 & 170 & 8 & 166.96 & 8.76 \\
        OpenAI GPT-4o & \cmark & 81.33 & 0.63 & 257 & 8 & 252.39 & 8.72 \\
        OpenAI o1-mini & \xmark & 50.82 & 4.82 & 127 & 719 & 124.82 & 705.00 \\
        OpenAI o1-mini & \cmark & 85.51 & 3.41 & 217 & 426 & 213.30 & 418.09 \\
        OpenAI o3-mini & \xmark & 59.69 & 9.65 & 166 & 977 & 163.04 & 957.92 \\
        OpenAI o3-mini & \cmark & 87.65 & 3.74 & 253 & 247 & 248.47 & 242.51 \\
        Mistral Large & \xmark & 4.59 & 0.49 & 198 & 18 & 194.44 & 17.86 \\
        Mistral Large & \cmark & 0.92 & 0.36 & 303 & 12 & 297.72 & 12.32 \\
        Llama 3.3 70B Instruct & \xmark & 44.49 & 0.32 & 284 & 9 & 278.92 & 9.04 \\
        Llama 3.3 70B Instruct & \cmark & 79.29 & 0.29 & 373 & 9 & 365.78 & 9.03 \\
        Google Gemma 3 27B & \xmark & 40.51 & 0.77 & 127 & 10 & 124.58 & 9.86 \\
        Google Gemma 3 27B & \cmark & 79.80 & 0.82 & 218 & 11 & 214.04 & 11.26 \\
        DeepSeek R1 & \xmark & 59.08 & 8.93 & 125 & 612 & 122.76 & 600.15 \\
        DeepSeek R1 & \cmark & 85.10 & 5.12 & 212 & 358 & 208.25 & 351.01 \\
        Qwen QwQ 32B & \xmark & 51.94 & 27.91 & 126 & 865 & 124.45 & 848.04 \\
        Qwen QwQ 32B & \cmark & 88.37 & 11.01 & 219 & 412 & 215.34 & 403.94 \\
        DeepSeek R1 Distill Qwen 32B & \xmark & 46.63 & 16.20 & 119 & 565 & 117.10 & 553.77 \\
        DeepSeek R1 Distill Qwen 32B & \cmark & 86.84 & 7.98 & 208 & 287 & 204.53 & 282.17 \\
      \bottomrule
    \end{tabular}%
  }
  \caption{Summary of tested models' performance on HotpotQA chemistry subset in terms of several evaluation metrics for both Contextual and Non-Contextual Setups}
  \label{tab:hotpotqa_results}
\end{table*}

\newpage
\subsection {A Multi-Hop QA Generation Example}
Figure \ref{fig:multihop} illustrates a typical multi-hop QA example derived from our knowledge-graph-based methodology. The context is drawn from chemical literature discussing the use of \textit{carbon dioxide} as a renewable feedstock for \textit{formic acid}, which then serves as a non-gaseous CO surrogate in \textit{carbonylation reactions}. By chaining these facts together, our approach constructs a question that requires integrating multiple pieces of information to arrive at the correct answer. This demonstrates how multi-hop reasoning, guided by entity relations and supplemented with descriptive metadata, enables more complex question generation and evaluation of large language models. Additionally, Figure~\ref{fig:qa_pipeline} shows the step-by-step process of deriving multi-hop questions from a knowledge graph, illustrating how entities, relations, and descriptive metadata are combined to construct more complex queries.

\begin{figure*}[hbpt]
\centering
\fbox{%
  \begin{minipage}{0.95\textwidth}
    \textbf{Context:}\\[6pt]
    [Source 1*]: \hlgreen{Carbonylation reactions} constitute a potent tool to manufacture carboxylic acids and 
    their derivatives both in industry and academic organic synthesis. In general, 
    carbonylation requires the use of toxic carbon monoxide, which thus usually demands 
    certified high-pressure reaction vessels. Therefore, developing non-gaseous CO surrogate 
    for conducting safe and facile-operation carbonylation is an important and ongoing 
    research topic. Among these established CO surrogates, \hlblue{formic acid} is one kind 
    of versatile atom.\\[6pt]

    [Source 2*]: The utilization of \hlred{carbon dioxide} as a C1 feedstock for the generation of 
    industrially relevant chemicals is also an interesting approach. CO\textsubscript{2} is 
    an attractive renewable C1 source, which can lead to \hlblue{formic acid}. Those 
    approaches would not only reduce carbon dioxide emissions through carbon capture but 
    also compensate sequestration costs by producing chemicals in global demand.\\[12pt]

    \textbf{Question:}\\
    What is the process that uses a substance, produced from \hlred{carbon dioxide} and 
    known as the simplest carboxylic acid with antibacterial and preservative properties, 
    as a non-gaseous surrogate to safely form carboxylic acids and their derivatives under 
    mild conditions?\\[6pt]

    \textbf{Answer:} \hlgreen{carbonylation reactions}\\[6pt]

    \textbf{Sentence-level supporting facts:}\\
    1) \hlblue{formic acid} can be produced from \hlred{carbon dioxide}.\\
    2) \hlblue{formic acid} is the simplest carboxylic acid with antibacterial and preservative properties.\\
    3) \hlblue{formic acid} can act as a non-gaseous CO surrogate.\\
    4) \hlgreen{carbonylation reactions} safely produce carboxylic acids under mild conditions 
       using \hlblue{formic acid} as a CO surrogate.\\[6pt]

    \textbf{Path (multi-hop chain of reasoning):}\\
    \hlred{carbon dioxide} \(\rightarrow\) \hlblue{formic acid} \(\rightarrow\) \hlgreen{carbonylation reactions}\\

    * \small{Source 1 and source 2 are coming from different documents. }
  \end{minipage}
}
\caption{An example of a multi-hop question-answer.}
\label{fig:multihop}
\end{figure*}

\begin{figure*}[hbpt]
\centering
\fbox{%
  \begin{minipage}{0.95\textwidth}
    \textbf{Context:}\\[6pt]
    [Source 1*]: 
    The most common way to functionalise
    two-dimensional materials such as \hlblue{graphene}
    is through reactions occurring in \hlred{solution}\,. \\[12pt]
    
    [Source 2*]: Radiolytic shielding via graphene
    \hlyellow{membranes} has been demonstrated,
    highlighting the role of the two-dimensional allotrope
    \hlblue{graphene}.\\[6pt]

    [Source 3*]:    A variety of chemisorptionbased solid sorbents exist, such as amines grafted onto porous solids like silica, cellulose, or metalorganic frameworks (MOFs). Membrane-based DAC captures CO$_2$ by selectively
    allowing CO$_2$ to permeate \hlyellow{membranes}
    while excluding other gases like \hlpurple{nitrogen}\,.\\[6pt]

    [Source 4*]: \hlgreen{Cr$_3$(Cr$_4$Cl)$_3$(BTT)$_8$$_2$
    (BTT$_3$ 1,3,5-benzenetristetrazolate)$_{12}$}\,
    MOF exhibits very high selectivity for O$_2$ over 
    \hlpurple{nitrogen}\,. While multiple factors can influence gas adsorption in MOFs, the origin of very high selectivity for O2 is perplexing because structurally similar metal ion MOFs are unselective.\\[6pt]

    \textbf{Question:}\\
    What is the metal–organic framework that exhibits very high oxygen
    selectivity over the major atmospheric diatomic gas, which is typically
    excluded by the selective barriers used for isolating carbon dioxide,
    and which act as radiolytic shields when formed from a two-dimensional
    carbon allotrope that is often functionalised in solution?\\[6pt]

    \textbf{Answer:} \hlgreen{Cr$_3$(Cr$_4$Cl)$_3$(BTT)$_8$$_2$
    (BTT$_3$ 1,3,5-benzenetristetrazolate)$_{12}$}\\[6pt]

    \textbf{Sentence-level supporting facts:}\\
    1) \hlred{Solution}-phase chemistry is the standard route to
       functionalise \hlblue{graphene}.\\
    2) \hlblue{Graphene} can form \hlyellow{membranes} that provide
       radiolytic shielding.\\
    3) \hlyellow{Membranes} used for DAC selectively exclude
       \hlpurple{nitrogen}.\\
    4) \hlgreen{Cr$_3$(Cr$_4$Cl)$_3$(BTT)$_8$$_2$} MOF shows very
       high O$_2$ selectivity over \hlpurple{nitrogen}.\\[6pt]

    \textbf{Path (multi-hop chain of reasoning):}\\
    \hlred{solution}
    $\rightarrow$ \hlblue{graphene}
    $\rightarrow$ \hlyellow{membranes}
    $\rightarrow$ \hlpurple{nitrogen}
    $\rightarrow$ \hlgreen{Cr$_3$(Cr$_4$Cl)$_3$(BTT)$_8$$_2$}\\[6pt]

    *\small{Sources 1–4 are extracted from four different documents.}
  \end{minipage}
  }
\caption{A 4-hop multi-document question–answer example.}
\label{fig:multihop-4hop}
\end{figure*}

\subsection{Impact of Context and Reasoning on output tokens count}

\begin{figure}[h]
    \centering
    \includegraphics[width=0.8\linewidth]{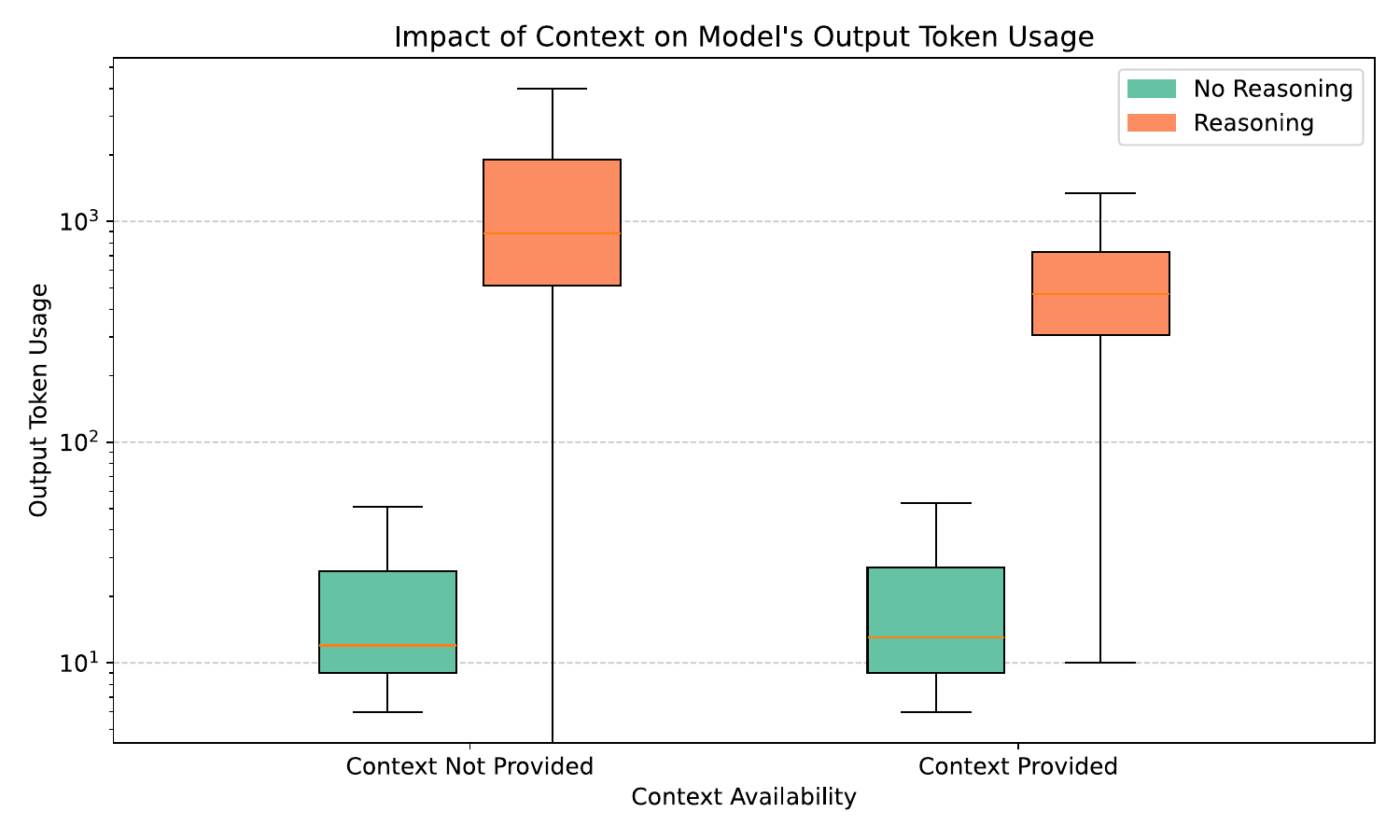}
    \caption{Visualization of the token usage distribution based on input context availability and model reasoning capability. The y-axis is log-scaled.}
    \label{fig:token_usage_context_reasoning}
\end{figure}

Figure~\ref{fig:token_usage_context_reasoning} illustrates the impact of context availability on the average number of output tokens generated by reasoning and non-reasoning models when answering questions. Non-reasoning models produce a similar number of tokens, as they do not engage in test-time reasoning. In contrast, for reasoning models, the number of tokens generated to answer questions decreases with the availability of context, indicating a potential requirement for less thinking when the context is available.

\newpage
\subsection{Performance Analysis Based on Number of Hops }

\begin{figure*}[ht!]
    \centering
    \includegraphics[width=\linewidth]{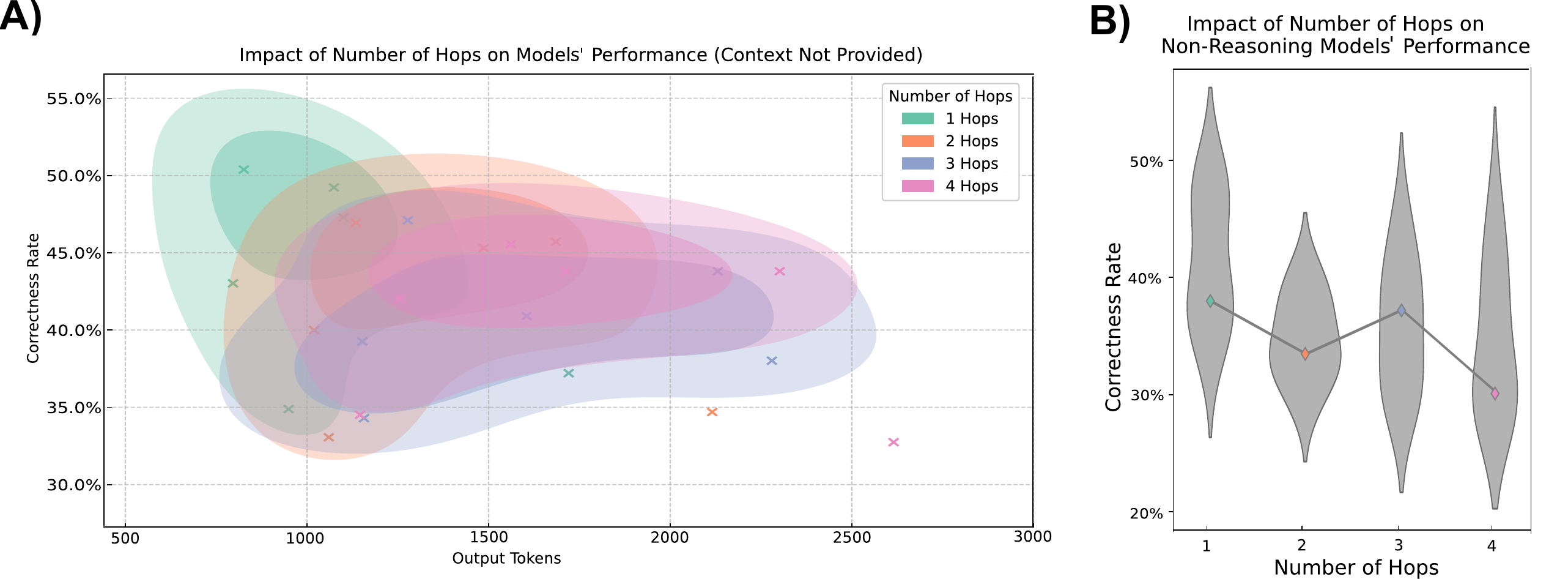}
    \caption{Analysis of the impact of the number of hops on the model's performance in the absence of context. A) Scatter and KDE plot of the number of hops vs. correctness rate and output tokens. B) Distribution of the non-reasoning models' performance for different numbers of hops.}
    \label{fig:hops_no_context}
\end{figure*}

\begin{figure*}[ht!]
    \centering
    \includegraphics[width=\linewidth]{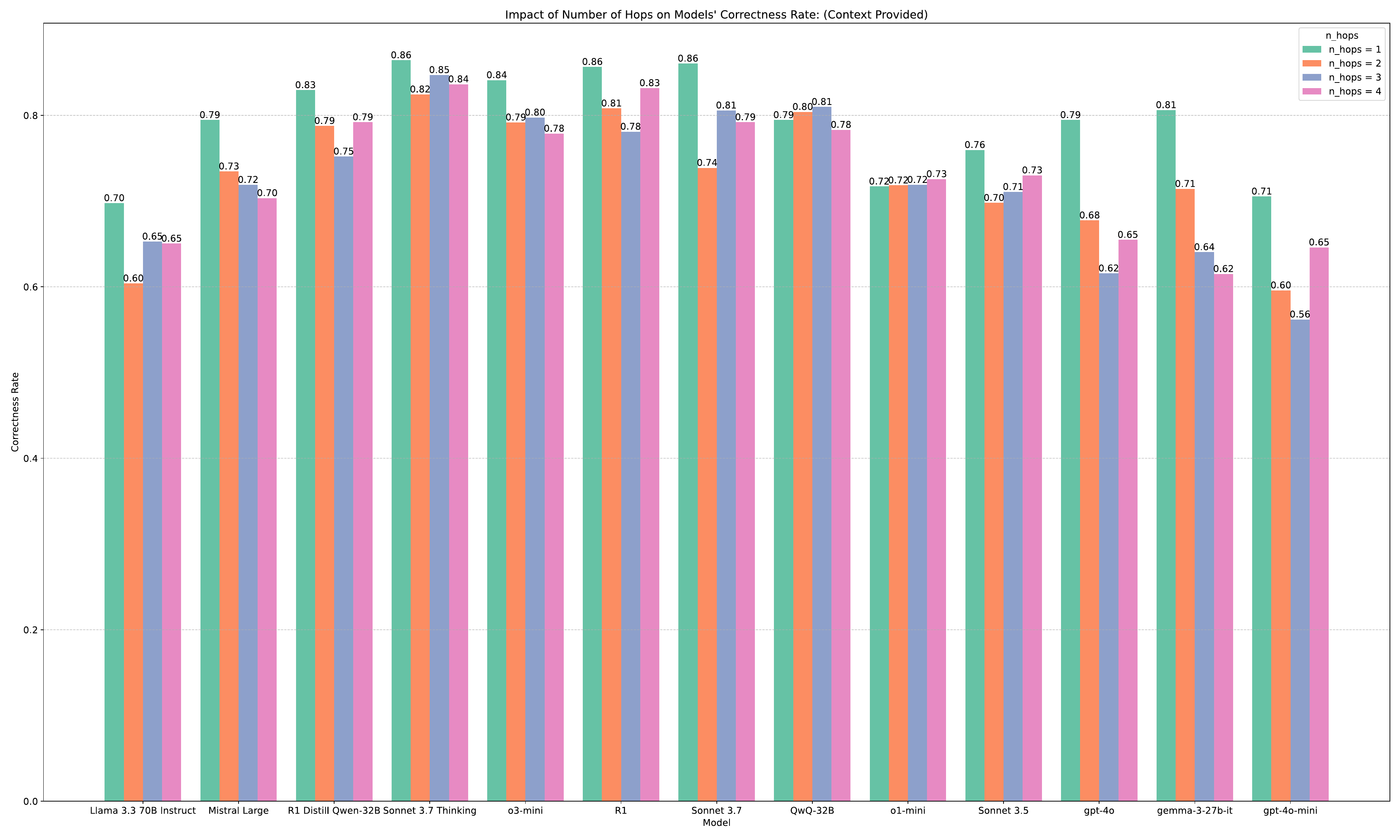}
    \caption{Overall performance of models as a function of the number of hops When context is provided.}
    \label{fig:performance_hops_with_context}
\end{figure*}

\begin{figure*}[ht!]
    \centering
    \includegraphics[width=\linewidth]{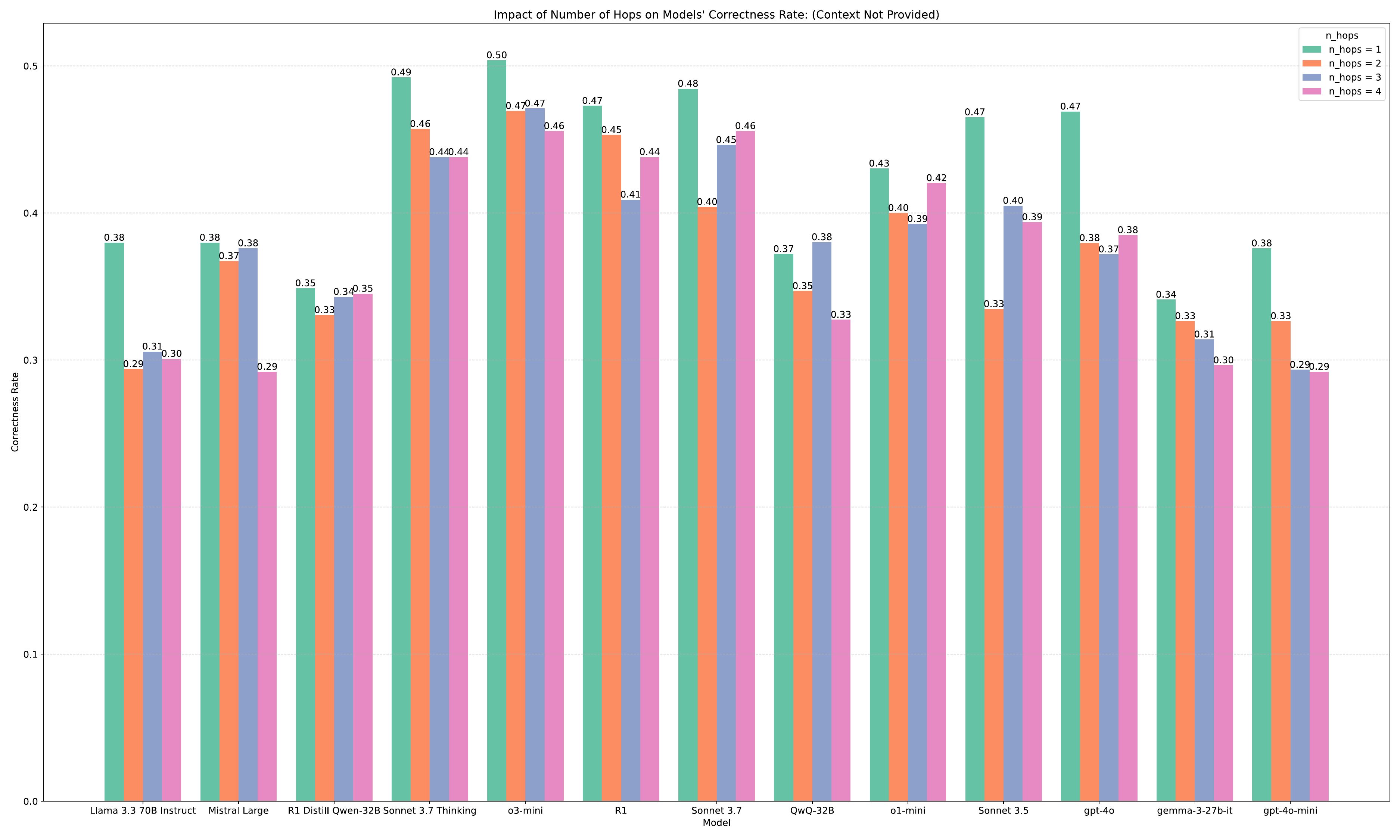}
    \caption{Overall performance of models as a function of the number of hops when context is not provided.}
    \label{fig:performance_hops_without_context}
\end{figure*}

\begin{figure*}[ht!]
    \centering
    \includegraphics[width=\linewidth]{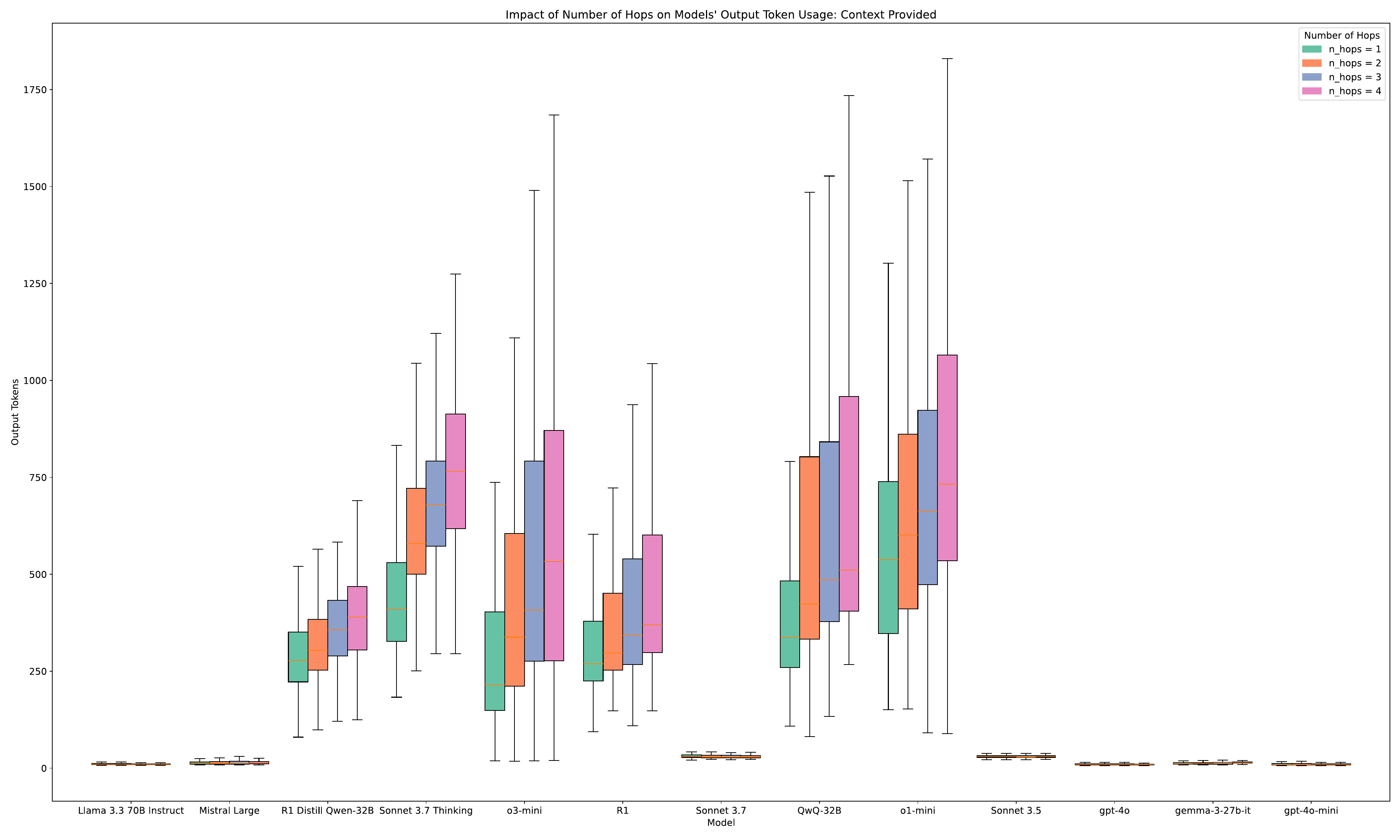}
    \caption{Token usage across different numbers of hops when context is provided.}
    \label{fig:token_hops_with_context}
\end{figure*}

\begin{figure*}[ht!]
    \centering
    \includegraphics[width=\linewidth]{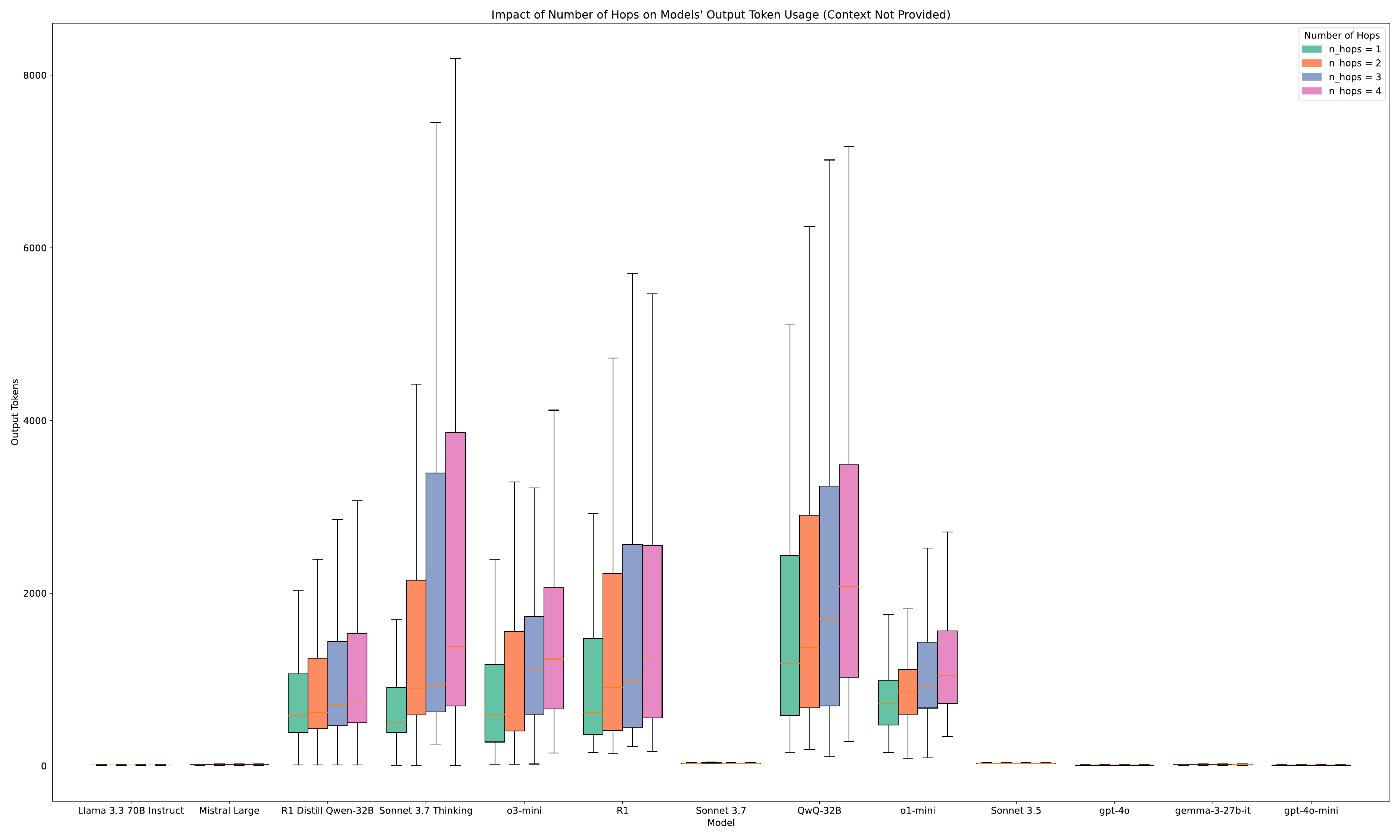}
    \caption{Token usage across different numbers of hops when context is not provided.}
    \label{fig:token_hops_without_context}
\end{figure*}

\begin{figure*}[ht!]
    \centering
    \includegraphics[width=\linewidth]{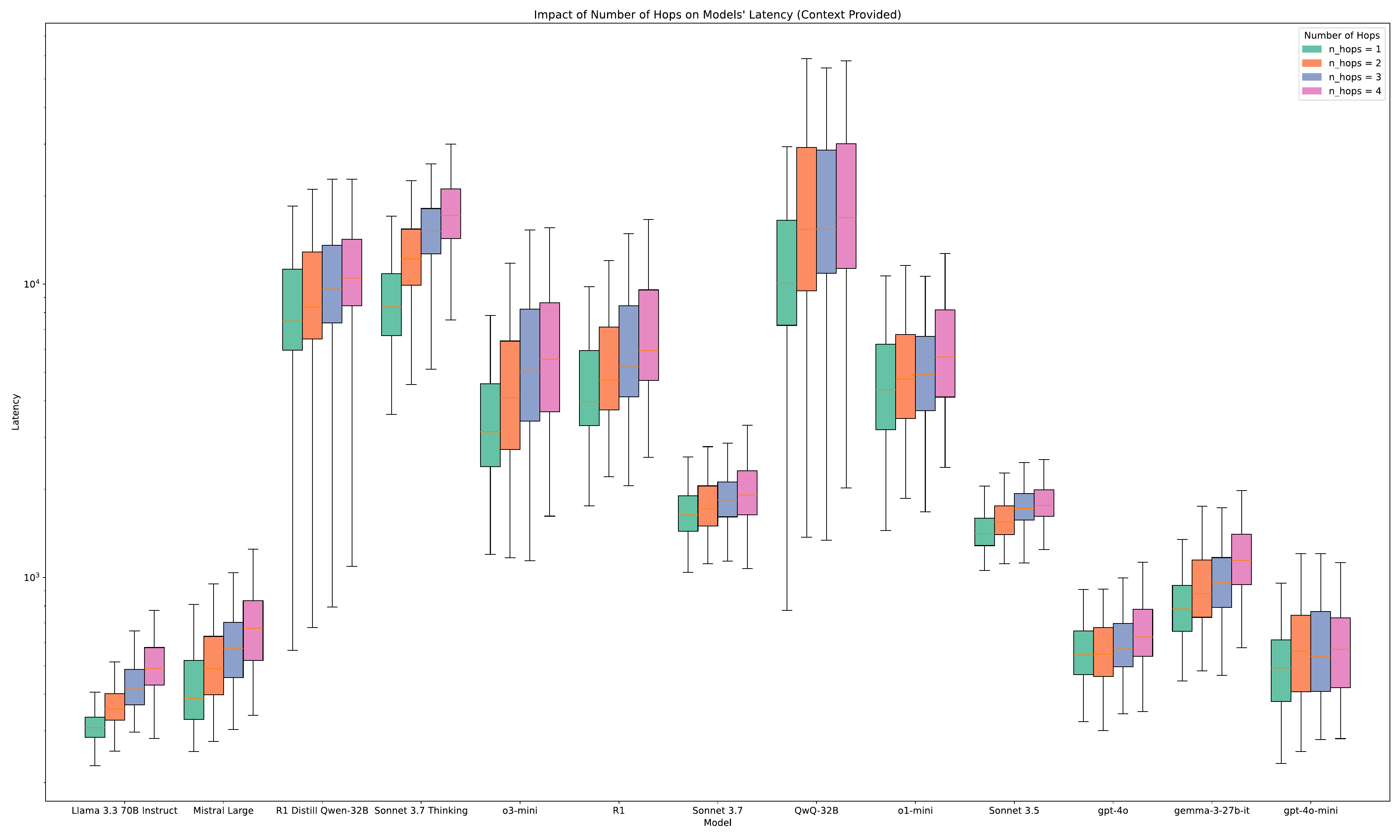}
    \caption{Latency measurements across different numbers of hops when context is provided.}
    \label{fig:latency_hops_with_context}
\end{figure*}

\begin{figure*}[ht!]
    \centering
    \includegraphics[width=\linewidth]{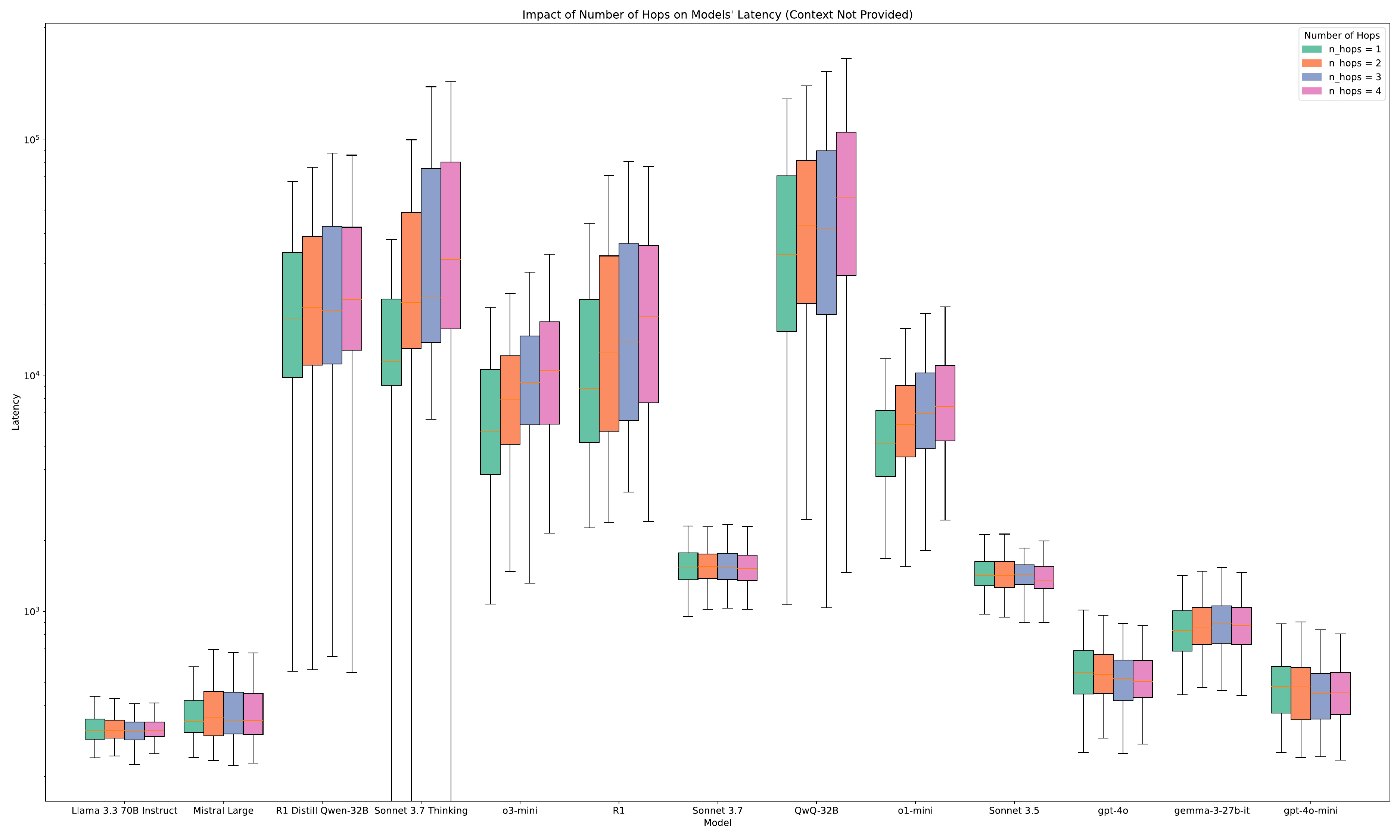}
    \caption{Latency measurements across different numbers of hops when context is not provided.}
    \label{fig:latency_hops_without_context}
\end{figure*}

Figure~\ref{fig:hops_no_context} illustrates how the number of hops affects performance in the absence of context. The data reveals that as the number of hops increases, the correctness rate declines while the number of output tokens rises. Additionally, both Figure~\ref{fig:hops_context} and Figure~\ref{fig:hops_no_context} show a negative correlation between token count and correctness when only one hop is used. This may suggest that overanalyzing simpler questions could lead to errors in the final answer. We visualized the overall performance of all models in Figure~\ref{fig:performance_hops_with_context}  and ~\ref{fig:performance_hops_without_context} to analyze how the correctness rate varies with the number of reasoning hops. In addition, token usage and latency metrics were separately depicted in Figures~\ref{fig:token_hops_with_context}, ~\ref{fig:token_hops_without_context},  ~\ref{fig:latency_hops_with_context} and~\ref{fig:latency_hops_without_context}, respectively, to provide a more detailed view of the efficiency and resource consumption as the reasoning depth increases.

The following figures illustrate the details of each model's performance, latency, and tokens used for question clusters requiring a different number of hops to be answered.

\end{document}